\documentclass[letterpaper,journal]{IEEEtran}

\usepackage{amsmath}
\usepackage{amssymb}
\usepackage{amsfonts}
\usepackage{booktabs}
\usepackage{tabularx}
\usepackage{algorithm}
\usepackage{algpseudocode}
\newcommand{\algcommentstyle}[1]{{\color{black!60}\itshape//~#1}}
\algrenewcommand{\algorithmiccomment}[1]{\hfill\algcommentstyle{#1}}
\makeatletter
\newcommand{\LineComment}[1]{\Statex \hskip\ALG@thistlm \algcommentstyle{#1}}
\makeatother
\usepackage{multirow}
\usepackage{graphicx}
\usepackage{subcaption}
\usepackage{overpic}
\usepackage[bookmarks=true,colorlinks]{hyperref}
\usepackage[table]{xcolor} %
\usepackage[utf8]{inputenc}
\usepackage[T1]{fontenc}
\usepackage[english]{babel}

\makeatletter
\let\NAT@parse\undefined
\makeatother
\usepackage[numbers,sort&compress]{natbib}

\newcommand{\bb}[1]{{\mathbb{#1}}}
\newcommand{\SO}[1]{\mathrm{SO}(#1)}
\newcommand{\SE}[1]{\mathrm{SE}(#1)}
\newcommand{\so}[1]{{\mathfrak{so}(#1)}}

\newcommand{\Exp}{\mathrm{Exp}}
\newcommand{\Log}{\mathrm{Log}}
\newcommand{\Ad}{\mathrm{Ad}}

\newcommand{\dexp}{\mathrm{dexp}}
\newcommand{\R}{\mathbb{R}}
\newcommand{\E}{\mathbb{E}}
\newcommand{\g}{\mathfrak{g}}
\newcommand{\sg}{\texttt{stopgrad}}
\renewcommand{\cal}{\mathcal}

\definecolor{linkcolor}{rgb}{0.0,0.0,1.0}
\definecolor{purduegold}{HTML}{C28E0E} %
\hypersetup{
bookmarksopen,
bookmarksnumbered,
colorlinks=true,
allcolors=purduegold,
}

\definecolor{hicell}{HTML}{ebd99f}  %

\newcommand{\stdv}[1]{\,\text{\scriptsize\ensuremath{\pm\,#1}}} %

\definecolor{grasp0}{rgb}{0.502,0.0,0.502}
\definecolor{grasp1}{rgb}{0.0,0.545,0.545}
\newcommand{\grenderwidth}{0.16\textwidth}

\newcommand{\bsrenderwidth}{0.128\textwidth}
\newcommand{\bsball}[1]{%
  \includegraphics[width=\bsrenderwidth,viewport=168 168 632 632,clip]{#1}}  %

\newcommand{\insertfig}{
    \includegraphics[width=0.8\linewidth,trim={2cm, 2cm, 2cm, 2cm},clip]{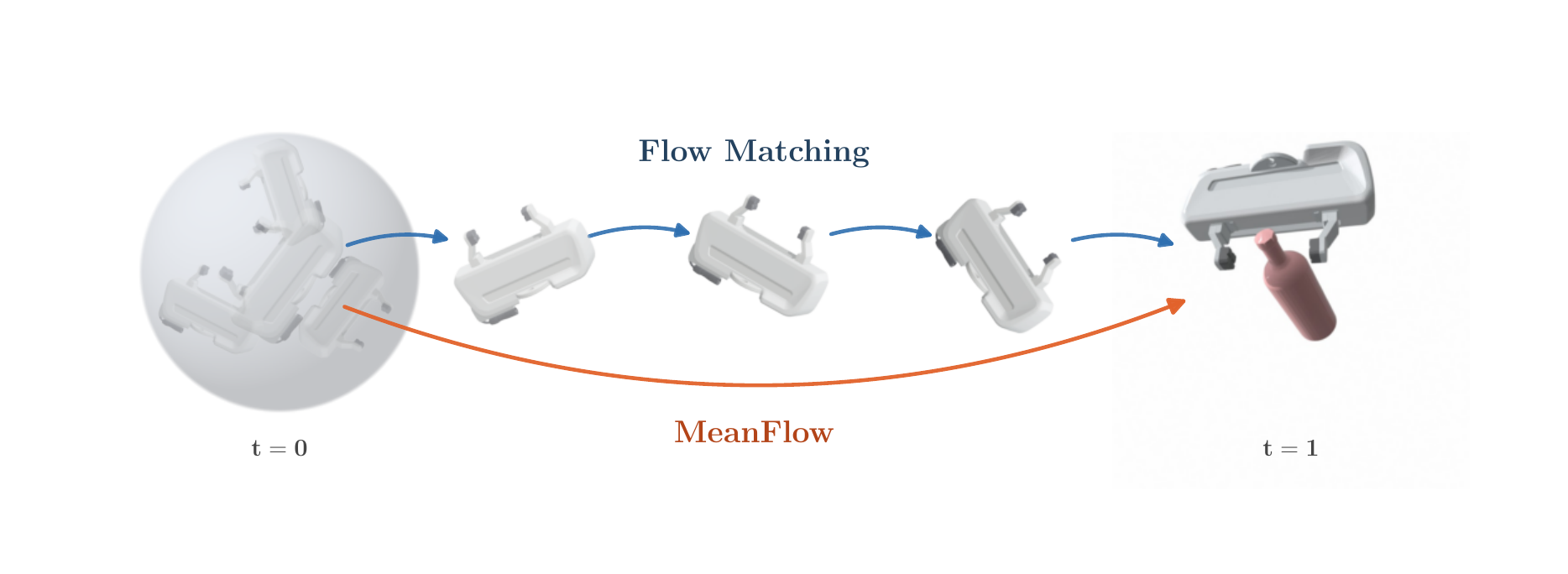}\captionof{figure}{\label{fig:teaser}
    Flow Matching iteratively moves grasp poses sampled from a prior distribution ($t=0$) towards the target distribution ($t=1$).
    By contrast, MeanFlow estimates the average velocity for a chord-form update, transporting grasp poses to the target distribution in substantially fewer steps.
    }
}
\makeatletter
\apptocmd{\@maketitle}{\centering\setcounter{figure}{0}\insertfig\vspace{-6mm}}{}{}%
\makeatother

\title{Fast Generative Grasping via Lie Group-Constrained MeanFlow}

\author{S. Talha Bukhari, Yi Wei, Ruiqi Ni, Zachary Kingston, and Aniket Bera%
\thanks{Authors are with the Department of Computer Science, Purdue University, West Lafayette, IN 47907, USA.
\texttt{\small \{bukhars, wei577@purdue.edu, ni117@purdue.edu, zkingston, aniketbera\}@purdue.edu}}
}

\begin{document}

\maketitle
\setcounter{figure}{1}  %
\markboth{Journal of \LaTeX\ Class Files,~Vol.~18, No.~9, September~2020}%
{How to Use the IEEEtran \LaTeX \ Templates}

\begin{abstract}
  Grasp synthesis is a core task in robotic manipulation, for which the solution typically forms a multimodal distribution rather than a point estimate.
  Generative robotic grasping aims to learn this distribution with deep generative models such as diffusion and flow-based approaches.
  The iterative nature of such generative models makes them flexible and generalizable; however, multi-step sampling impedes the time-critical operation required in robotics.
  We devise an approach to fast generative grasping based on MeanFlow on the product Lie group $\cal{G} = \SO{3} \times \R^3$.
  The training objective couples a purely algebraic semigroup consistency condition with Riemannian Conditional Flow Matching on $\cal{G}$ that anchors the average velocity to the data distribution.
  The resulting Lie Group-constrained MeanFlow formulation samples reliable grasps in $\leq 5$ network evaluations, matching the grasp generation performance of state-of-the-art diffusion and flow-based models on the ACRONYM dataset at millisecond-scale inference latency (up to $39\times$ speed-up).
  We further demonstrate that the approach directly translates to real-world robotic grasping without additional training or domain adaptation, exhibiting robust grasp synthesis under observation noise.
\end{abstract}


\section{Introduction}\label{sec:introduction}

Robotic grasping entails mapping a perceived object representation to an end-effector pose that achieves a stable grasp.
For a given object, the set of stable grasps forms a multimodal distribution in $\SE{3}$, motivating the learning of object-conditioned grasp distributions over point estimation~\citep{Eppner2021ACRONYM}.
Generative models trained on grasp datasets capture this multimodal structure, and, conditioned on point cloud observations, sample multiple feasible grasps~\citep{Urain2023SE3DiffusionFields, Lim2024EquiGraspFlow, Chen2024BRIDGER} while also admitting composition with downstream objectives such as collision avoidance and reachability~\citep{Urain2023SE3DiffusionFields}.

Diffusion~\citep{Ho2020DDPM, Song2021Score} and Flow Matching~\citep{Lipman2023FlowMatching} are the prevailing generative modeling approaches in robotic manipulation~\citep{Chi2023DiffusionPolicy, Ze2024DP3, Hu2024AdaFlow, Black2024Pi0}.
Their iterative generation procedure enables broad mode coverage and stable training; the step-by-step refinement progressively transports samples onto the data manifold which, in the case of robotic grasping, encodes the implicit contact constraints of valid grasps.
This fidelity comes at a cost: inference requires tens to hundreds of integration steps, precluding the sampling rates that reactive, closed-loop execution demands.

Recent work on few-step generation aims to reduce the number of network function evaluations (NFEs) required for sampling.
Consistency models~\citep{Song2023Consistency} and shortcut models~\citep{Frans2024Shortcut} accelerate sampling but optimize self-referential targets rather than aligning with the underlying transport vector field.
MeanFlow~\citep{Geng2025MeanFlow} instead predicts the \emph{average velocity} over a time interval, a well-defined property of the probability path.
The resulting objective is teacher-free and simulation-free, its optimum is characterized independently of the network, and the learned average velocity supports one-step and few-step generation.
\citet{Woo2026RiemannianMeanFlow} extend the framework to Riemannian manifolds through an algebraic semigroup consistency objective that avoids high-variance differential terms for stable training, and apply it to protein backbone generation.

Reactive, closed-loop execution in robotics demands sampling rates that few-step generation can supply, motivating the use of MeanFlow.
While MeanFlow formulations have been adapted to robotics tasks such as manipulation policy learning~\citep{Sheng2025MP1}, manifold-constrained formulations are relatively less explored, and whether few-step sampling on a manifold preserves the implicit contact constraints that iterative diffusion and flow models resolve through many refinement steps lacks empirical evaluation backed by physics-based simulation and real-world experimentation.

Towards this objective, we employ MeanFlow on the product Lie group $\cal{G} = \SO{3} \times \R^3$ for fast grasp pose generation, which we call \emph{GraspMF}.
The approach couples the algebraic semigroup identity with the Conditional Flow Matching anchor on $\cal{G}$.
Consequently, the average velocity learned by MeanFlow amounts to an implicit, learned integration of the underlying velocity field over the sampling interval, so that a few-step jump along the predicted geodesic chord reaches the contact-feasible grasp manifold while avoiding the large-step truncation error that degrades diffusion and flow-based samplers at low step counts.
On the ACRONYM dataset, GraspMF matches the grasp success rate and distribution coverage of state-of-the-art diffusion and flow-based methods at millisecond-scale latency and remains stable across sampling budgets.
We further demonstrate that GraspMF can handle real-world grasping with observation noise while maintaining efficient inference.

\section{Related Work}\label{sec:related_work}

\subsection{Generative Models on Manifolds}\label{sec:related_manifold}

Score-based generative models and continuous normalizing flows have been extended to Riemannian manifolds by replacing Euclidean drift and diffusion terms with their manifold-valued counterparts.
Riemannian score-based generative modeling defines forward and reverse stochastic differential equations driven by Brownian motion on the manifold and a manifold-aware score~\citep{DeBortoli2022RiemannianScore}.
Riemannian Flow Matching~\citep{Chen2024RiemannianFM} reformulates conditional flow matching with geodesic interpolants and a velocity field tangent to the manifold, removing the need for divergence computation or trajectory simulation during training.
When the manifold is a Lie group, this tangent-field representation simplifies: left- or right-trivialization identifies its tangent bundle with the product of the group and its Lie algebra, so the velocity field can be learned as a Lie-algebra-valued map~\citep{Yim2023SE3Diffusion, Sola2018LieTheory}.
In the protein-frame setting, $\SE{3}$ flow matching instantiates this construction with the product structure $\SO{3} \times \R^3$ and geodesic conditional paths~\citep{Yim2023SE3FlowMatching, Bose2024FoldFlow}.

Consistency models~\citep{Song2023Consistency}, rectified flow~\citep{Liu2023RectifiedFlow}, progressive distillation~\citep{Salimans2022ProgressiveDistillation}, and shortcut models~\citep{Frans2024Shortcut} accelerate diffusion or flow-based inference in Euclidean spaces by enforcing self-consistency along trajectories or by training shorter-step samplers from longer-step teachers.
MeanFlow~\citep{Geng2025MeanFlow} reformulates this through the average velocity over a time interval, yielding a simulation-free training objective without a teacher; SplitMeanFlow~\citep{Guo2025SplitMeanFlow} derives an equivalent algebraic interval-splitting consistency identity that removes the Jacobian--vector product appearing in the original MeanFlow loss.
Riemannian MeanFlow~\citep{Woo2026RiemannianMeanFlow} establishes three equivalent characterizations of the average velocity on a manifold (Eulerian, Lagrangian, semigroup), where the algebraic semigroup objective scales reliably in high dimensions, with application to protein-frame $\SE{3}^N$ generation.
We formulate the algebraic semigroup identity as a consistency objective for $\SE{3}$ grasp poses, coupled with a Riemannian Conditional Flow Matching anchor that ties the average velocity to the data distribution.

\subsection{Generative Grasping}\label{sec:related_grasping}

Early learning-based grasp synthesis approaches regressed grasp parameters directly, or scored sampled candidates from depth or point-cloud observations with discriminative models~\citep{Mahler2019DexNet, Sundermeyer2021ContactGraspNet}, with limited handling of the multimodality of the grasp set.
Generative approaches replace deterministic prediction with sampling from a learned grasp distribution, the earliest approach being conditional variational autoencoders that generate and rank candidate grasps~\citep{Mousavian20196DOFGraspNet}.

$\SE{3}$-DiffusionFields~\citep{Urain2023SE3DiffusionFields} learns a smooth grasp-energy field on $\SE{3}$ via denoising score matching, by perturbing grasp poses with Lie-algebraic Gaussian noise injected through the exponential map, and samples grasps by running Langevin dynamics on the learned energy field.
EquiGraspFlow~\citep{Lim2024EquiGraspFlow} casts grasp generation as $\SE{3}$-equivariant flow matching, where deterministic integration of the learned Lie-algebra-valued vector field requires fewer steps than denoising diffusion.
Furthermore, its equivariant architecture guarantees by construction that generated grasps transform consistently with rigid transformations of the object.
BRIDG{\scriptsize{E}}R~\citep{Chen2024BRIDGER} instead reduces the sampling budget by replacing the uninformative prior with an informative source distribution under the stochastic interpolants framework~\citep{Albergo2023StochasticInterpolants}.
GraspLDM~\citep{Barad2024GraspLDM} places the diffusion process in the latent space of a variational autoencoder and synthesizes $\SE{3}$ grasps by decoding the latent samples.
\citet{Carvalho2024GraspDiffusionNetwork} adopt a DDPM formulation on $\SO{3}\times\R^3$ for grasp generation from partial point clouds.
The grasp generation quality of these methods degrades at low sampling budgets, whereas our proposal retains the fidelity of iterative refinement at five or fewer NFEs.
Concurrent work by~\citet{Zhong2026RMFManifolds} also applies a MeanFlow formulation to grasp generation by relating the average velocity to the instantaneous velocity via a covariant derivative along the path, and formulating a decomposition of the training objective.
However, the gradients of the decomposed objective can conflict, for which the work resorts to multi-task learning, and computing the covariant derivative can hinder training stability while incurring significant computational overhead.
Furthermore, an extensive evaluation against prior state-of-the-art is required, and, as we show in Sec.~\ref{sec:exp_ablation}, the decomposed MeanFlow objective does not yield competitive performance against prior diffusion and flow-based methods.

A separate line of work emphasizes the interdependence of grasp synthesis and geometry reconstruction.
\citet{Bukhari2025VariationalShapeInference} improve grasp generation through a shape-inference stage that learns geometric features from implicit neural representations, prioritizing grasp quality under imperfect and partial observations rather than sampling efficiency.
\citet{Song2024ImplicitGraspDiffusion} condition grasp diffusion on local geometric features from implicit neural representations, unifying dense prediction with sampling-based generation.
We likewise couple grasp generation with geometric supervision through an auxiliary shape reconstruction objective that grounds the MeanFlow formulation on the object geometry.

\section{Preliminaries}\label{sec:preliminaries}

\begin{figure*}[t]
  \centering
  \setlength{\tabcolsep}{-2pt}
  \begin{tabular}{@{}c@{\hspace{4pt}}|@{\hspace{4pt}}c@{}}
    \begin{tabular}{@{}ccccc@{}}
      \multicolumn{5}{c}{\footnotesize $\SO{3}$ Flow Matching}\\[1pt]
      \bsball{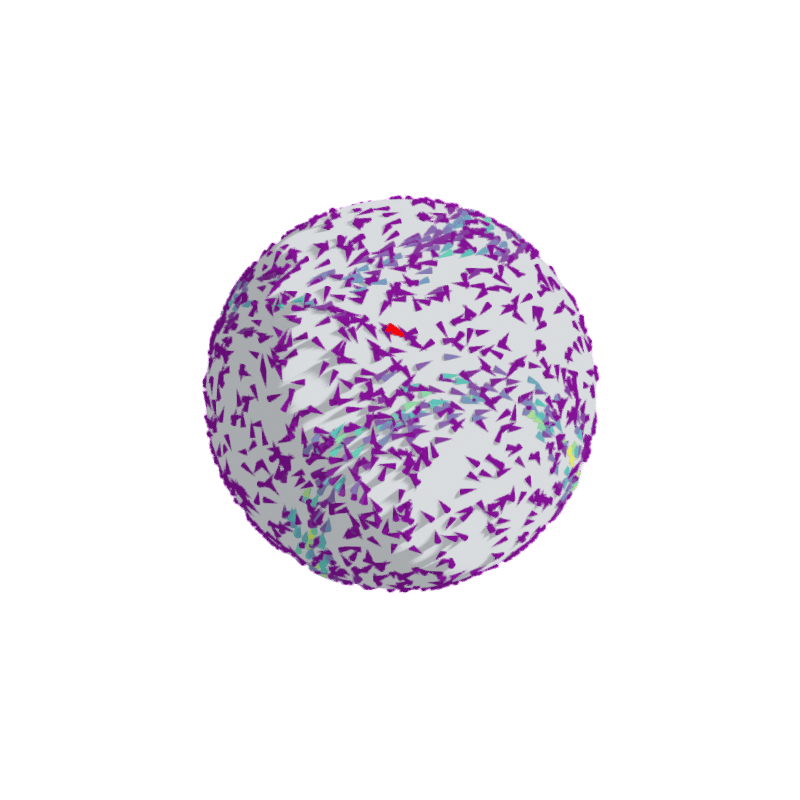} &
      \bsball{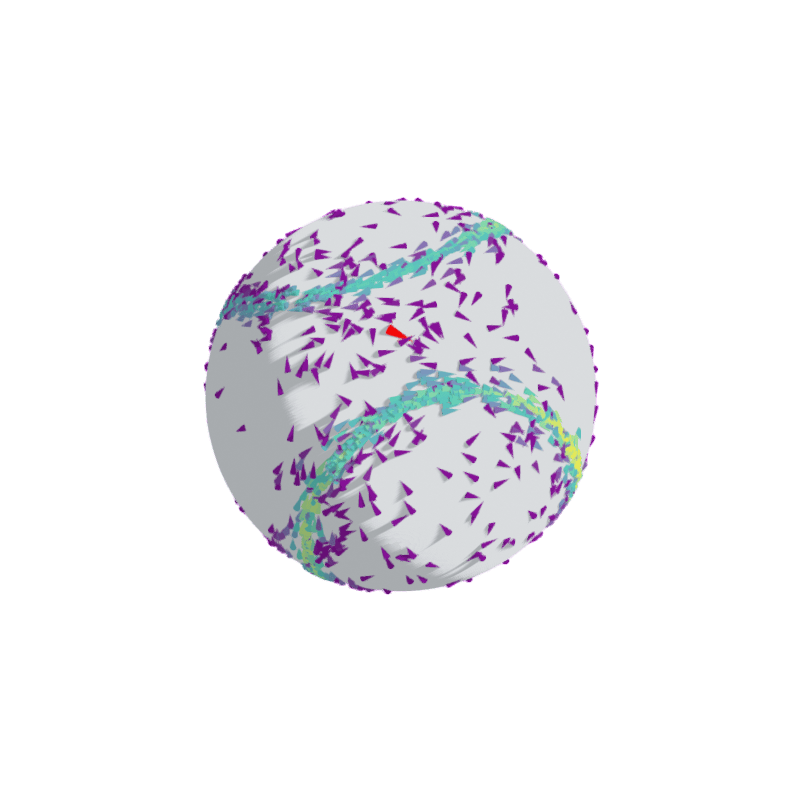} &
      \bsball{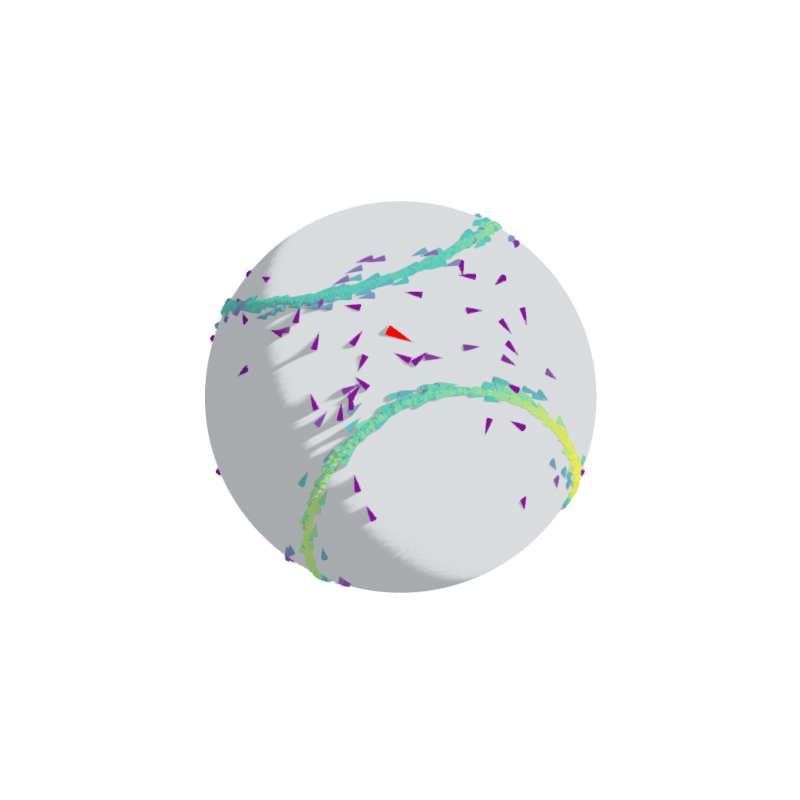} &
      \bsball{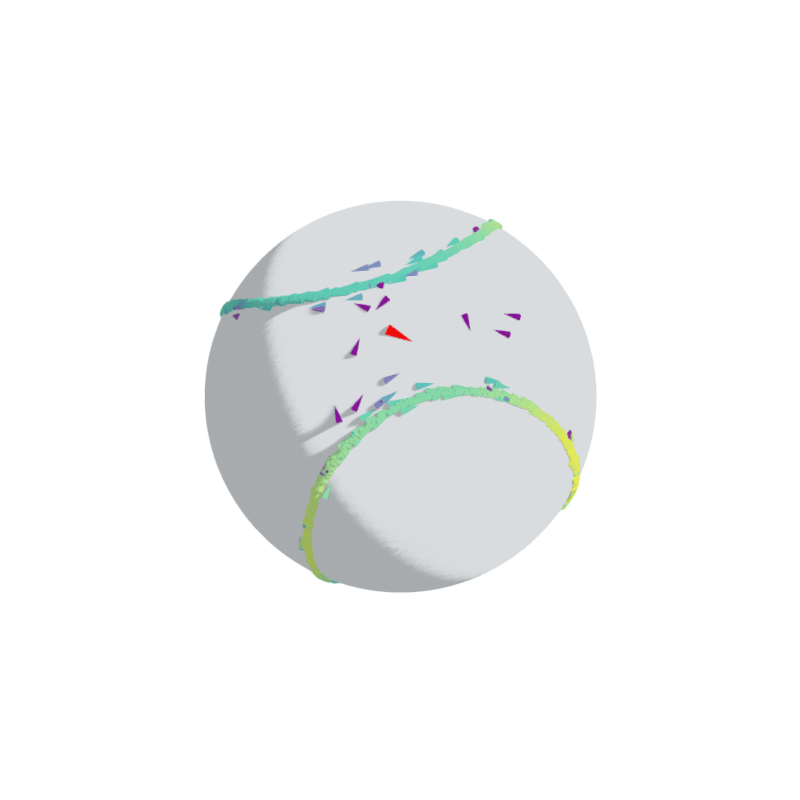} &
      \bsball{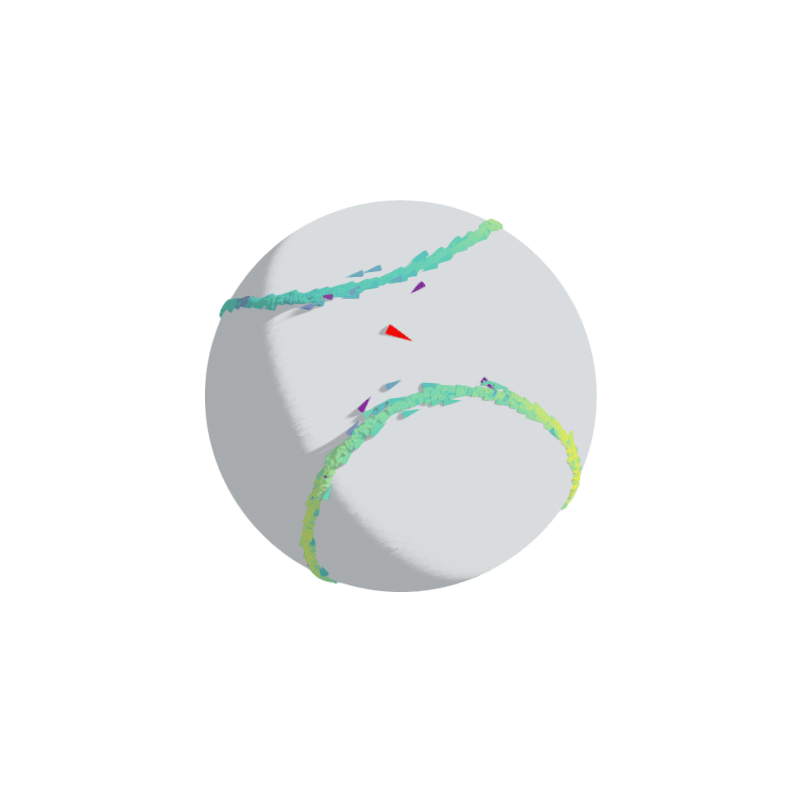}\\[1pt]
      {\footnotesize $T = 1$} & {\footnotesize $T = 2$} & {\footnotesize $T = 5$} & {\footnotesize $T = 10$} & {\footnotesize $T = 50$}
    \end{tabular}
    &
    \begin{tabular}{@{}ccc@{}}
      \multicolumn{3}{c}{\footnotesize $\SO{3}$ MeanFlow}\\[1pt]
      \bsball{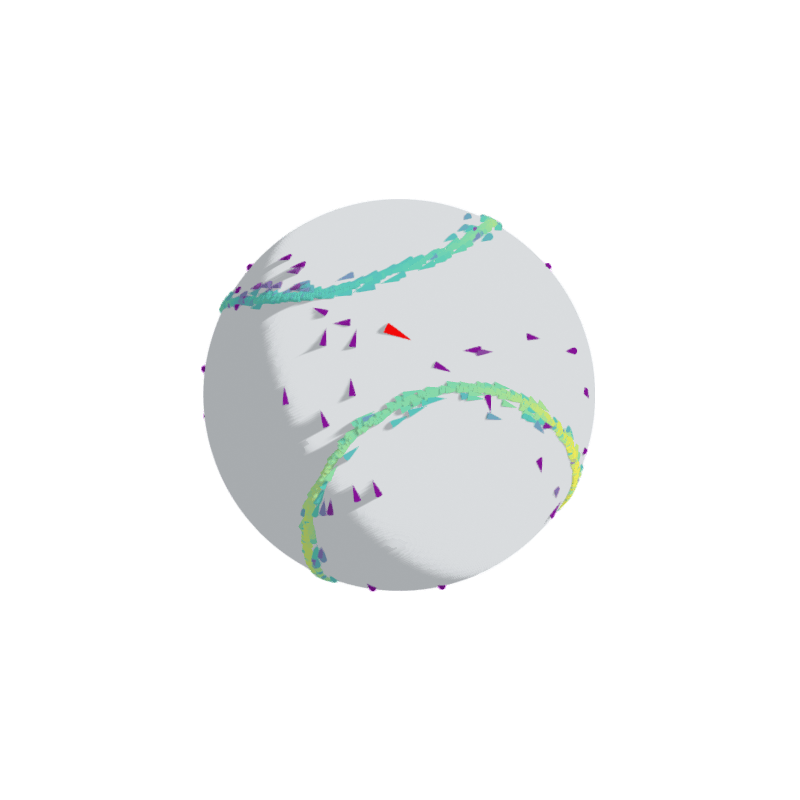} &
      \bsball{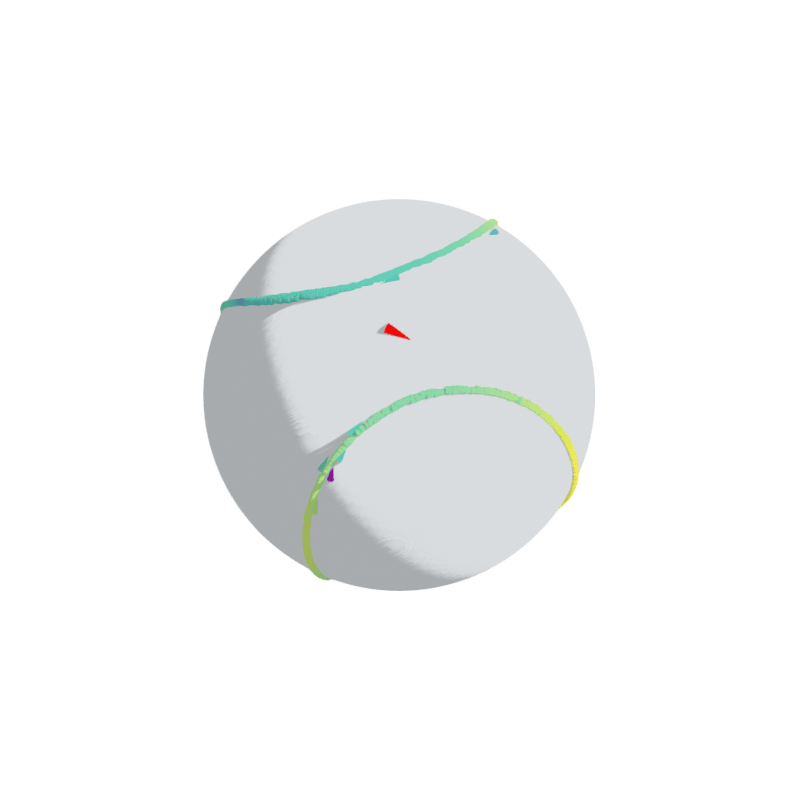} &
      \bsball{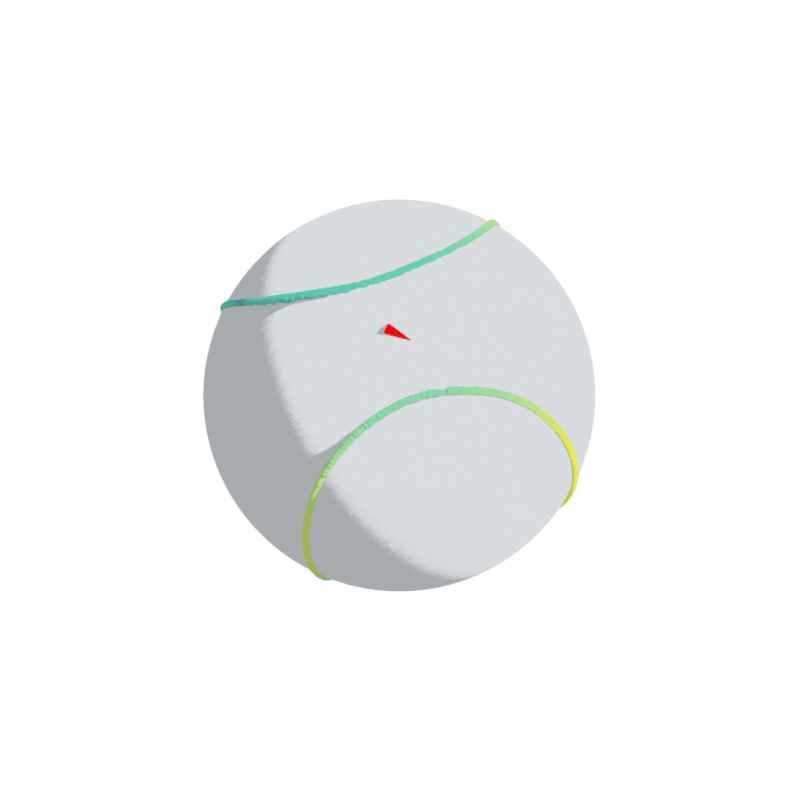}\\[1pt]
      {\footnotesize $T = 1$} & {\footnotesize $T = 2$} & {\footnotesize $T = 3$}
    \end{tabular}
  \end{tabular}
  \caption{A visual comparison of generative modeling on the $\SO{3}$ matrix Lie group.
  We design a target distribution (of elements of $\SO{3}$) whose support is a closed curve on the unit sphere $S^2$, traced by the $z$-axis of the frame with the $y$-axis aligned to the tangent of the curve.
  Each sampled rotation $R$ is rendered as a cone placed at $R e_3$ and oriented along $R e_2$, with $\{e_1, e_2, e_3\}$ the standard basis of $\R^3$.
  The cone is colored by a kernel density estimate of the target evaluated at that sample (yellow: high density, purple: low density), and the red cone marks the identity rotation $I$.
  Flow Matching (\emph{left}) leaves a large fraction of samples off the seam at low step counts and requires tens of integration steps to concentrate on the support, whereas MeanFlow (\emph{right}) recovers the seam within three steps.}\label{fig:ball-seam}
\end{figure*}

\subsection{Riemannian Manifolds and Lie Groups}\label{sec:prelim_geom}

Informally, an $n$-dimensional manifold $\cal{M}$ is a topological space that locally resembles (i.e., is \emph{homeomorphic} to) Euclidean space $\bb{R}^n$.
A manifold is smooth or differentiable if it additionally has $C^\infty$ differential structure.
At a point $x \in \cal{M}$, the tangent space $T_x \cal{M}$ is the vector space comprising vectors tangent to smooth curves through $x$.
The disjoint union of all tangent spaces forms the tangent bundle $T\cal{M} = \bigsqcup_{x \in \cal{M}} T_x \cal{M}$.
A Riemannian manifold $(\cal{M}, g)$ is additionally equipped with a Riemannian metric $g$, a smoothly varying inner product on the tangent spaces $g_x: T_x \cal{M} \times T_x \cal{M} \to \R$.
A Lie group $\cal{G}$ is a group that is also a differentiable manifold, in which multiplication $\circ: \cal{G} \times \cal{G} \to \cal{G}$ and inversion $(\cdot)^{-1}: \cal{G} \to \cal{G}$ are smooth maps.
The left action of $\cal{G}$ on itself, $L_x: y \mapsto x \circ y$ for $x, y \in \cal{G}$, is a diffeomorphism and identifies the tangent space at $x$ with the tangent space at the identity $e \in \cal{G}$ (i.e., the Lie algebra $\g$) through its differential $(L_x)_*: T_e \cal{G} \to T_x \cal{G}$.
The Lie algebra is equipped with an alternating bilinear map called the Lie bracket $[\cdot, \cdot]: \g \times \g \to \g$.
Elements of the Lie algebra map to the Lie group via the group exponential $\exp_{\cal{G}}: \g \to \cal{G}$, a diffeomorphism near the identity whose local inverse is the group logarithm $\log_{\cal{G}}$.
Matrix Lie groups are Lie groups whose elements are matrices, with matrix multiplication as the group operation; the matrix exponential $\exp$ and the principal matrix logarithm $\log$ realize $\exp_{\cal{G}}$ and $\log_{\cal{G}}$.

Orientation-preserving rigid-body transformations in three dimensions form the matrix Lie group $\SE{3} = \{(R, p) \mid R \in \SO{3}, p \in \R^3\}$, where $\SO{3}$ is the special orthogonal group of rotations.
As a Lie group, $\SE{3} = \SO{3} \ltimes \R^3$ is a semidirect product and admits no bi-invariant Riemannian metric~\citep{Milnor1976LeftInvariant, Park1995ScrewMetric}.
Prior work~\citep{Yim2023SE3Diffusion, Yim2023SE3FlowMatching, Urain2023SE3DiffusionFields, Lim2024EquiGraspFlow} therefore models on the direct product group $\cal{G} = \SO{3} \times \R^3$, which is diffeomorphic to $\SE{3}$ and admits a bi-invariant metric.
This metric is induced by the $\Ad$-invariant inner product on the Lie algebra $\g = \so{3} \oplus \R^3$,
\begin{equation}
  \label{eq:g_metric}
  \langle (A, u), (B, v) \rangle_{\g} \;=\; \langle A, B \rangle_{\so{3}} + u^\top v,
\end{equation}
where $A, B \in \so{3}$, $u, v \in \R^3$, and the $\Ad$-invariant inner product on $\so{3}$~\citep{Hall2015LieGroups} is
\begin{equation}
  \label{eq:so3_metric}
  \langle A, B \rangle_{\so{3}} \;=\; \tfrac{1}{2}\mathrm{tr}(A B^\top) \;=\; -\tfrac{1}{2}\mathrm{tr}(A B).
\end{equation}
Elements of $\so{3}$ are skew-symmetric matrices, identified with $\R^3$ by the linear isomorphisms $(\cdot)^\wedge: \R^3 \to \so{3}$ and $(\cdot)^\vee: \so{3} \to \R^3$, under which~\eqref{eq:so3_metric} reads $\langle A, B \rangle_{\so{3}} = (A^\vee)^\top B^\vee$.
Being block-diagonal, the induced product metric $g_{\cal{G}} = g_{\SO{3}} \oplus g_{\R^3}$ decouples the rotational and translational components of a pose.
We write $\|X\|_x$ for the norm of $X \in T_x \cal{G}$ under $g_{\cal{G}}$, and $\|\cdot\|_{\g}$ for the norm induced by~\eqref{eq:g_metric} on the Lie algebra; left-invariance gives $\|X\|_x = \|(L_{x^{-1}})_* X\|_{\g}$.
In this work, we model grasp poses as rigid-body transforms $H = (R, p)$, with $R \in \SO{3}$ the gripper orientation and $p \in \R^3$ the gripper position, and learn their distribution on $\cal{G}$.

\subsection{Riemannian Conditional Flow Matching on $\cal{G}$}\label{sec:prelim_cfm}

Let $\rho_0$ be a tractable prior on $\cal{G}$, and let $\rho_1$ be a target distribution conditioned on an observation $c \in \cal{C}$.
The conditioning $c$ is shared by $\rho_1$, the conditional and marginal velocities, and every learned map introduced below; we state this dependence here once and suppress $c$ from the notation thereafter.
Riemannian Conditional Flow Matching~\citep{Chen2024RiemannianFM, Tong2024CFM} defines, for each pair $(H_0, H_1) \sim \rho_0 \otimes \rho_1$ whose rotational parts are non-antipodal (so that $\Log_{H_0}(H_1)$ is well-defined), the geodesic interpolant
\begin{equation}
  \label{eq:geodesic_interp}
  H_t \;=\; \Exp_{H_0}\big( t \cdot \Log_{H_0}(H_1) \big), \qquad t \in [0,1],
\end{equation}
and the conditional velocity at $H_t$,
\begin{equation}
  \label{eq:cfm_velocity}
  v_t(H_t \mid H_0, H_1) \;=\; \frac{1}{1-t}\,\Log_{H_t}(H_1) \;\in\; T_{H_t}\cal{G},
\end{equation}
which is tangent to $\cal{G}$ and generates~\eqref{eq:geodesic_interp} as the integral curve of $\dot{H}_t = v_t(H_t \mid H_0, H_1)$ starting at $H_0$. %
A vector field $v_\theta: \cal{G} \times [0,1] \to T\cal{G}$, with $v_\theta(H,t) \in T_H \cal{G}$, is trained to match the marginal velocity by minimizing
\begin{equation}
  \label{eq:cfm_loss}
  \mathcal{L}_{\mathrm{CFM}}(\theta) \;=\; \E_{\substack{(H_0, H_1) \\ t \sim \cal{U}[0,1]}}\!\left[\big\| v_\theta(H_t, t) - v_t(H_t \mid H_0, H_1) \big\|^2_{H_t}\right].
\end{equation}
At inference the ODE $\dot{H}_t = v_\theta(H_t, t)$ is integrated on $\cal{G}$ from $H_0 \sim \rho_0$ to $t = 1$ via the exponential map at each step, yielding a sample from $\rho_1$.

\section{Riemannian MeanFlow on $\cal{G}$}\label{sec:method}

We employ the Riemannian MeanFlow formulation~\citep{Woo2026RiemannianMeanFlow} on the product Lie group $\cal{G} = \SO{3} \times \R^3$, equipped with the bi-invariant metric of Sec.~\ref{sec:prelim_geom}.
We adopt the \emph{endpoint} ($H_1$) parameterization, in which the network predicts the clean grasp directly and the average velocity and flow map used for training and inference are obtained from this prediction in closed form.
This parameterization inherits the stability of endpoint regression from Flow Matching and the two-time structure of flow maps, and has been reported to outperform direct average-velocity prediction~\citep{Woo2026RiemannianMeanFlow}.
Throughout this section we use the left-trivialization of Sec.~\ref{sec:prelim_geom}: tangent vectors are carried to the Lie algebra $\g$, so that residuals are compared in a single base-point-independent inner-product space.
For a marginal velocity field $v: \cal{G} \times [0,1] \to T\cal{G}$ generating the probability path between $\rho_0$ and $\rho_1$, we write its left-trivialized representation as
\begin{equation}
  \label{eq:xi_def}
  \xi(H, t) \;:=\; H^{-1}\, v(H, t) \;\in\; \g,
\end{equation}
so that $v(H, t) = (L_H)_* \xi(H, t)$ and the integral curves of $v$ satisfy $\dot{H}_t = (L_{H_t})_* \xi(H_t, t)$, which we abbreviate as $\dot H_t = H_t \xi(H_t, t)$.
The integral curves of $v$ define the marginal flow map $\Phi_{s,t}: \cal{G} \to \cal{G}$, which pushes the marginal density at time $s$ forward to that at time $t$.
Whenever the chord $H^{-1}\,\Phi_{s,t}(H)$ lies within the injectivity radius of $\exp_\cal{G}$, the flow map determines the chord-form \emph{average velocity} $u: \cal{G} \times \Delta \to \g$ defined on the simplex of ordered times $\Delta := \{(s,t) \in [0,1]^2 : s \le t\}$:
\begin{equation}
  \label{eq:avg_velocity}
  u(H, s, t) \;:=\; \frac{1}{t-s}\,\log_\cal{G}\!\big( H^{-1}\, \Phi_{s,t}(H) \big) \;\in\; \g, \; s < t,
\end{equation}
extended to the time-diagonal by the limit $u(H, s, s) = \xi(H, s)$.
MeanFlow trains a network to estimate $u$, so that sampling evaluates the flow map directly instead of integrating $\xi$.

\subsection{Endpoint Parameterization and the Induced Flow Map}\label{sec:method_avg_velocity}

The network $X_\theta: \cal{G} \times \Delta \to \cal{G}$ maps a state $H$ at time $s$ and an interval endpoint $t$ to an endpoint prediction,
\begin{equation}
  \label{eq:x1_pred}
  \hat{H}_1 \;=\; X_\theta(H, s, t) \;=\; (\hat R_1, \hat p_1) \;\in\; \cal{G}.
\end{equation}
The prediction is interpreted as the clean grasp obtained by transporting $H$ at the average velocity of the time interval $[s, t]$ over the remaining horizon $[s, 1]$.
Inverting this interpretation recovers the model's estimate $\bar{u}_\theta$ of the average velocity~\eqref{eq:avg_velocity},
\begin{equation}
  \label{eq:trivialized_avg_def}
  \bar{u}_\theta(H, s, t) \;:=\; \frac{1}{1-s}\,\log_\cal{G}\!\big(H^{-1}\, X_\theta(H, s, t)\big) \;\in\; \g, \; s < 1,
\end{equation}
together with the induced flow map $\Phi_\theta$ that transports $H$ from time $s$ to time $t$,
\begin{align}
  \label{eq:flow_map}
  \Phi_\theta(H, s, t) &\;:=\; H \cdot \exp_\cal{G}\!\big((t-s)\, \bar{u}_\theta(H, s, t)\big) \\
                       &\;=\; \Exp_H\!\Big(\tfrac{t-s}{1-s}\,\Log_H(\hat H_1)\Big).
\end{align}
The flow map~\eqref{eq:flow_map} returns the point at geodesic parameter $(t-s)/(1-s)$ along the minimizing geodesic from $H$ to the predicted endpoint $\hat H_1$.
By the direct product metric, the flow map decouples into a rotational geodesic and a translational secant,
\begin{align}
  \label{eq:flow_map_factors}
  \Phi_\theta(H, s, t)_R &\;=\; R\,\exp\!\Big(\tfrac{t-s}{1-s}\,\log(R^\top \hat R_1)\Big), \\ %
  \Phi_\theta(H, s, t)_p &\;=\; p + \tfrac{t-s}{1-s}\,(\hat p_1 - p).
\end{align}
Here, two boundary cases anchor the construction:
\begin{itemize}
  \item At the diagonal $t = s$, $\bar u_\theta(H, s, s)$ reduces to the model's instantaneous left-trivialized velocity at $H$, the counterpart of the diagonal limit of~\eqref{eq:avg_velocity}; Riemannian Conditional Flow Matching matches it with the marginal velocity, grounding the formulation in the data distribution,
  \begin{equation}
    \label{eq:boundary}
    \bar u_\theta(H, s, s) \;=\; \frac{1}{1-s}\,\log_\cal{G}\!\big(H^{-1} X_\theta(H, s, s)\big) \;{\triangleq}\; \xi(H, s).
  \end{equation}
  \item At the full horizon $t = 1$, the fraction in~\eqref{eq:flow_map} is unity and $\Phi_\theta(H, s, 1) = X_\theta(H, s, 1)$ is the single-step grasp prediction.
\end{itemize}
Hence, both the average velocity~\eqref{eq:trivialized_avg_def} and the induced map~\eqref{eq:flow_map} exclude the corner $(s, t) = (1, 1)$, the only point of $\Delta$ at which the remaining horizon $1 - s$ vanishes.

\subsection{Semigroup (Flow-Map) Consistency on $\cal{G}$}\label{sec:method_semigroup}

The exact flow map satisfies the semigroup property $\Phi_{s,t} = \Phi_{r,t} \circ \Phi_{s,r}$ for any $0 \le s \le r \le t \le 1$, since flowing from time $s$ to time $t$ factors through every intermediate time $r$.
We require the induced map $\Phi_\theta$ to obey the same identity,
\begin{equation}
  \label{eq:semigroup_map}
  \Phi_\theta(H_s, s, t) \;=\; \Phi_\theta\!\big( \Phi_\theta(H_s, s, r),\, r,\, t \big),
\end{equation}
i.e.\ a single chord-form jump over $[s, t]$ agrees with the two-step composition through the intermediate time $r$.
Whenever the chord segments involved lie within the injectivity radius of $\exp_\cal{G}$, we obtain the equivalent velocity-space identity in the left-trivialized tangent space at $H_s$,
\begin{equation}
  \label{eq:semigroup_log}
  (t-s)\, \bar{u}_\theta(H_s, s, t)
  \;=\;
  \log_\cal{G}\!\Big( H_s^{-1}\, \Phi_\theta\!\big( \Phi_\theta(H_s, s, r),\, r,\, t \big) \Big).
\end{equation}
Importantly, this identity is purely algebraic: it involves only forward evaluations of $X_\theta$, the group exponential $\exp_\cal{G}$, and the group logarithm $\log_\cal{G}$, and contains no covariant derivative, no $\dexp^{-1}$, and no partial derivative of the network.
This is the property exploited by~\citet{Guo2025SplitMeanFlow} in the Euclidean setting and by~\citet{Woo2026RiemannianMeanFlow} on Riemannian manifolds; both report that the algebraic semigroup objective scales more stably than the differential (Eulerian/Lagrangian) identities, whose targets contain curvature-dependent variance from derivative-of-network terms.

Let us denote the two-step composition by
\begin{equation}
  \label{eq:tilde_gs}
  \tilde H_r \;:=\; \Phi_\theta(H_s, s, r), \qquad \tilde H_t \;:=\; \Phi_\theta(\tilde H_r, r, t),
\end{equation}
where $\tilde H_r = (\tilde R_r, \tilde p_r)$, $\tilde H_t = (\tilde R_t, \tilde p_t)$, and $\hat H_1 = X_\theta(H_s, s, t)$.
Then, the direct product structure splits the semigroup identity~\eqref{eq:semigroup_log} into one identity per factor:
\begin{align}
  \label{eq:semigroup_factors}
  \tfrac{t-s}{1-s}\,\log\!\big(R_s^\top \hat R_1\big) &\;=\; \log\!\big( (R_s^\top \tilde R_r)(\tilde R_r^\top \tilde R_t) \big), \\
  \tfrac{t-s}{1-s}\,\big(\hat p_1 - p_s\big) &\;=\; (\tilde p_r - p_s) + (\tilde p_t - \tilde p_r).
\end{align}
With~\eqref{eq:flow_map_factors}, dividing the translational identity by $t-s$ gives the Euclidean interval split of~\citet{Guo2025SplitMeanFlow}.
Near the identity, the Baker--Campbell--Hausdorff formula gives
\begin{equation}
  \log \left(\exp A \exp B \right) = A + B + \tfrac{1}{2}[A, B] + \cdots \;,
\end{equation}
for $A := \log(R_s^\top \tilde R_r)$ and $B := \log(\tilde R_r^\top \tilde R_t)$.
The same split therefore holds on the rotational factor only up to the bracket terms, which vanish exactly when $A$ and $B$ commute, i.e., when the two chords turn about a common axis.

\begin{algorithm}[t]
  \caption{Training MeanFlow on $\cal{G}$.}\label{alg:training}
  \begin{algorithmic}[1]
    \Repeat
    \State sample $H_1 \sim p_{\mathrm{data}}$;\;\, draw $H_0^{\mathrm{cfm}}, H_0^{\mathrm{semi}} \sim \rho_0$

    \LineComment{Flow Matching anchor~\eqref{eq:cfm_anchor}}
    \State sample anchor time $t_{\mathrm{cfm}}$
    \State $H_{t_{\mathrm{cfm}}} \gets \Exp_{H_0^{\mathrm{cfm}}}\!\big( t_{\mathrm{cfm}}\, \Log_{H_0^{\mathrm{cfm}}}(H_1) \big)$
    \State $\hat H_1 \gets X_\theta(H_{t_{\mathrm{cfm}}}, t_{\mathrm{cfm}}, t_{\mathrm{cfm}})$ \Comment{endpoint estimate}
    \State $\hat\eta \gets \log_\cal{G}\!\big(H_{t_{\mathrm{cfm}}}^{-1}\hat H_1\big)$;\;\, $\eta \gets \log_\cal{G}\!\big(H_{t_{\mathrm{cfm}}}^{-1} H_1\big)$
    \State $\ell_{\mathrm{CFM}} \gets \big\| \tfrac{1}{1-t_{\mathrm{cfm}}}\,(\hat\eta - \eta) \big\|^2_\g$

    \LineComment{Semigroup Consistency~\eqref{eq:semi_loss}}
    \State sample $(s, r, t)$ with $0 \le s \le r \le t \le 1$, $s < t$
    \State $H_s \gets \Exp_{H_0^{\mathrm{semi}}}\!\big( s\, \Log_{H_0^{\mathrm{semi}}}(H_1) \big)$
    \State $\tilde H_r \gets \Phi_\theta(H_s, s, r)$;\;\, $\tilde H_t \gets \Phi_\theta(\tilde H_r, r, t)$
    \State $\tilde u(s, t) \gets \tfrac{1}{t-s}\,\log_\cal{G}\!\big( H_s^{-1}\, \tilde H_t \big)$ \Comment{consistency target}
    \State $\ell_{\mathrm{semi}} \gets w(r)\,\big\| \bar u_\theta(H_s, s, t) - \sg(\tilde u(s, t)) \big\|^2_\g$

    \LineComment{Combined objective~\eqref{eq:rmf_loss}}
    \State $\ell \gets \lambda_{\mathrm{cfm}}\, \ell_{\mathrm{CFM}} + \lambda_{\mathrm{semi}}\, \ell_{\mathrm{semi}}$
    \State Take gradient descent step on $\nabla_\theta \ell$
    \Until{converged}
  \end{algorithmic}
\end{algorithm}

\subsection{Training Objective}\label{sec:method_train}

The network is the endpoint predictor $X_\theta$ of~\eqref{eq:x1_pred}, with output in $\cal{G}$; the average velocity $\bar u_\theta$ and flow map $\Phi_\theta$ are the derived quantities~\eqref{eq:trivialized_avg_def}--\eqref{eq:flow_map}.
The semigroup identity~\eqref{eq:semigroup_map} is data-free: the degenerate predictor $X_\theta(H, s, t) \equiv H$ (zero displacement, $\bar u_\theta \equiv 0$) satisfies it identically.
Therefore, an explicit single-time anchor against the data distribution is required.
The resulting anchored objective is consistent: the exact average velocity is a global minimizer, and, under regularity conditions, any idealized optimum reproduces the exact flow map, by the Riemannian counterpart~\citep{Woo2026RiemannianMeanFlow} of the interval-splitting theorem of~\citet{Guo2025SplitMeanFlow}.
Here we detail the two terms of the training objective, summarized in Algorithm~\ref{alg:training}.

\emph{Flow-Matching anchor.}
At the diagonal $s = t$ we regress the predicted endpoint onto the data sample through the conditional flow velocity defined on $\mathcal{G}$~\eqref{eq:cfm_velocity},
\begin{align}
  \label{eq:cfm_anchor}
  \mathcal{L}_{\mathrm{CFM}}(\theta) &\;=\; \E\!\left[ \big\| \bar{u}_\theta(H_t, t, t) - \xi_t^{\mathrm{cfm}} \big\|^2_\g \right], \\
  \bar{u}_\theta(H_t, t, t) &\;=\; \tfrac{1}{1-t}\,\log_\cal{G}\!\big(H_t^{-1} \hat H_1\big), \nonumber \\
  \xi_t^{\mathrm{cfm}} &\;=\; \tfrac{1}{1-t}\,\log_\cal{G}\!\big(H_t^{-1} H_1\big), \nonumber \\
  \hat H_1 &\;=\; X_\theta(H_t, t, t), \nonumber
\end{align}
where $H_t$ is the geodesic interpolant~\eqref{eq:geodesic_interp} of a sampled pair $(H_0, H_1)$, $\bar{u}_\theta(H_t, t, t)$ is the model average velocity~\eqref{eq:trivialized_avg_def} at the time diagonal, and $\xi_t^{\mathrm{cfm}} := H_t^{-1} v_t(H_t \mid H_0, H_1)$ is the left-trivialization of the conditional velocity~\eqref{eq:cfm_velocity}.
Equivalently,~\eqref{eq:cfm_anchor} regresses the predicted endpoint onto the data grasp in the left-trivialized tangent space at $H_t$, reweighted by $1/(1-t)$.
On the factors the residual is $\tfrac{1}{1-t}\big(\log(R_t^\top \hat R_1) - \log(R_t^\top R_1)\big)$ in rotation and $\tfrac{1}{1-t}(\hat p_1 - p_1)$ in translation.
By the standard marginal--conditional argument~\citep{Lipman2023FlowMatching, Chen2024RiemannianFM}, $\E[\xi_t^{\mathrm{cfm}} \mid H_t] = \xi(H_t, t)$, so the minimizer of~\eqref{eq:cfm_anchor} satisfies the boundary condition~\eqref{eq:boundary} with respect to the marginal velocity.

\emph{Semigroup loss.}
The base state $H_s$ is the geodesic interpolant~\eqref{eq:geodesic_interp} at time $s$, and $\tilde H_r, \tilde H_t$ are the two-step states~\eqref{eq:tilde_gs}.
Treating the composed two-step state of~\eqref{eq:semigroup_log} as a self-consistency target via a stop-gradient~\citep{Geng2025MeanFlow, Woo2026RiemannianMeanFlow}, the semigroup loss compares the single-step average velocity against the chord velocity of the composition,
\begin{align}
  \label{eq:semi_loss}
  \tilde u(s, t) &\;=\; \tfrac{1}{t-s}\log_\cal{G}\!\big( H_s^{-1} \tilde H_t \big), \nonumber \\
  \mathcal{L}_{\mathrm{semi}}(\theta) &\;=\; \E\left[ w(r) \big\| \bar u_\theta(H_s, s, t) - \sg(\tilde u(s, t)) \big\|^2_\g \right].
\end{align}
Here $\tilde u(s, t) \in \g$ is the chord velocity of the frozen two-step composition and $\sg(\cdot)$ denotes the stop-gradient operator; the intermediate states $\tilde H_r, \tilde H_t$ are formed under $\sg$ and the target is treated as a constant during the training step.
The expectation is over $H_0 \sim \rho_0$, $H_1 \sim p_{\mathrm{data}}$, and time triplets $(s, r, t)$ with $0 \le s \le r \le t \le 1$ and $s < t$.
The scalar $w(r)$ down-weights $r \to 1$, where the average velocity over the sub-interval $[r, t]$ carries a $1/(1-r)$ factor that amplifies prediction noise.

\emph{Combined objective.}
The full training objective is the weighted sum
\begin{equation}
  \label{eq:rmf_loss}
  \mathcal{L}_{\mathrm{RMF}}(\theta) \;=\; \lambda_{\mathrm{cfm}}\, \mathcal{L}_{\mathrm{CFM}}(\theta) \;+\; \lambda_{\mathrm{semi}}\, \mathcal{L}_{\mathrm{semi}}(\theta),
\end{equation}
with weights $\lambda_{\mathrm{cfm}}, \lambda_{\mathrm{semi}} > 0$.
At $\lambda_{\mathrm{semi}} = 0$ the model reduces to plain Riemannian CFM endpoint regression.

\subsection{Inference}\label{sec:method_inference}

For a partition $0 = t_0 < t_1 < \cdots < t_T = 1$, where $T$ is the sampling budget (the number of integration steps), a grasp is generated by drawing $H_0 \sim \rho_0$ and iterating the induced flow map $\Phi_\theta$~\eqref{eq:flow_map},
\begin{equation}
  \label{eq:inference_iter}
  H_{t_{k+1}} \;=\; \Phi_\theta(H_{t_k}, t_k, t_{k+1}), \qquad k = 0, \ldots, T-1.
\end{equation}
For $T = 1$ the geodesic fraction in~\eqref{eq:flow_map} is unity and a single network call returns the grasp directly, $H_1 = X_\theta(H_0, 0, 1)$.
The iteration is explicit: $X_\theta$ is evaluated at the left endpoint of each sub-interval, so each step costs exactly one network evaluation plus one closed-form $\exp$/$\log$ pair.
Each update is a chord-form jump toward the estimated clean grasp rather than an Euler step of a learned vector field: whenever $\Phi_\theta$ satisfies the semigroup identity~\eqref{eq:semigroup_map}, the $T$-step composition returns the single-step prediction $X_\theta(H_0, 0, 1)$ for every partition.

The iteration recovers a Lie--Euler scheme for the trivialized ODE $\dot H_t = H_t\, \xi(H_t, t)$~\citep{Iserles2000LieGroupMethods} in the fine-partition limit, by the continuity of $X_\theta$ in its second time argument together with the boundary condition~\eqref{eq:boundary},
\begin{equation}
  \label{eq:diagonal_limit}
  \lim_{\Delta t \to 0} \bar u_\theta(H_{t_k}, t_k, t_k + \Delta t) \;=\; \xi(H_{t_k}, t_k),
\end{equation}
but differs from it at any fixed step size $\Delta t$.
A Lie--Euler step freezes the velocity at $\xi(H_{t_k}, t_k)$ across $[t_k, t_k + \Delta t]$ and incurs a truncation error in $\cal{O}(\Delta t^2)$, whereas $\bar u_\theta(H_{t_k}, t_k, t_k + \Delta t)$ targets the constant trivialized velocity whose displacement over that interval reproduces the exact flow map~\eqref{eq:avg_velocity}.
The residual error of~\eqref{eq:inference_iter} is therefore not a truncation error in the numerical sense but the approximation error $\bar u_\theta - u$ of the two-time field.
Hence, the inference procedure exchanges one long-interval prediction for a composition of short-interval predictions; we support this notion empirically in Sec.~\ref{sec:exp_budget}.
Each update requires the principal logarithm $\log_\cal{G}\!\big(H_{t_k}^{-1}\, X_\theta(H_{t_k}, t_k, t_{k+1})\big)$, which is unique unless the rotational chord is antipodal, a measure-zero configuration handled by the stable $\SO{3}$ logarithm branch of Sec.~\ref{sec:implementation}.

\begin{table}[t]
  \centering\vspace{2mm}
  \small
  \setlength{\tabcolsep}{4pt}
  \resizebox{\linewidth}{!}{%
  \begin{tabular}{lcccc}
    \toprule
    \multirow{2}{*}{Method} & \multicolumn{2}{c}{SR $(\%)$ $\uparrow$} & \multicolumn{2}{c}{EMD $\downarrow$} \\
    \cmidrule(lr){2-3} \cmidrule(lr){4-5}
    & ID & OOD & ID & OOD \\
    \midrule
    SE3Dif~\citep{Urain2023SE3DiffusionFields}
    & $70.40\stdv{18.37}$ & $57.60\stdv{20.84}$ & $0.4649\stdv{0.1705}$ & $0.4880\stdv{0.1009}$ \\
    BRIDG{\scriptsize{E}}R~\citep{Chen2024BRIDGER}
    & $59.38\stdv{34.23}$ & $63.77\stdv{29.41}$ & $0.6057\stdv{0.1849}$ & $0.5589\stdv{0.1390}$ \\
    EGF~\citep{Lim2024EquiGraspFlow}
    & $\underline{85.48}\stdv{22.77}$ & $65.09\stdv{23.97}$ & $0.3866\stdv{0.1008}$ & $0.4805\stdv{0.1125}$ \\
    VSIGD~\citep{Bukhari2025VariationalShapeInference}
    & $68.57\stdv{20.20}$ & $\underline{69.05}\stdv{20.00}$ & $0.4966\stdv{0.1692}$ & $0.4675\stdv{0.0963}$ \\
    \midrule
    \textbf{GraspMF} ($T = 5$)
    & $\mathbf{87.40}\stdv{17.43}$ & $\mathbf{71.73}\stdv{25.57}$ & $\underline{0.3702}\stdv{0.0942}$ & $\mathbf{0.4191}\stdv{0.1108}$ \\
    \textbf{GraspMF} ($T = 1$)
    & $81.11\stdv{15.91}$ & $66.34\stdv{25.70}$ & $\mathbf{0.3698}\stdv{0.1039}$ & $\underline{0.4218}\stdv{0.1065}$ \\
    \bottomrule
  \end{tabular}%
  }
  \vspace{1mm}
  \caption{Grasping performance on objects from the ACRONYM dataset:
  grasp success rate (SR) and coverage (EMD).
  Each comparison method is reported at the sampling budget of its original formulation, and GraspMF at a low-step ($T = 5$) and single-step ($T = 1$) setting.
  SR and EMD are each subdivided into in-domain (ID), evaluated on held-out instances of the training shape categories, and out-of-domain (OOD), evaluated on instances from disjoint shape categories.
  The inference cost of the same settings is reported in Table~\ref{tab:latency_nfe}.
  \textbf{Best} and \underline{second-best} metrics are highlighted.}\label{tab:main}
\end{table}

\subsection{Implementation Details}\label{sec:implementation}

\emph{Network.}
For our proposal, we adopt the lightweight neural network architecture of SE3Dif~\citep{Urain2023SE3DiffusionFields}, replacing the scalar energy head with a pose prediction head.
A VNN point cloud encoder~\citep{deng2021vector} maps the object point cloud $c \in \R^{N \times 3}$ ($N = 1024$) to a descriptor $z$ that conditions the network.
A grasp pose $H$ is mapped to a high-dimensional representation via a fixed set of keypoints that rigidly transform with $H$, which are then mapped to embeddings.
The two time arguments $(s, t)$ are each encoded with random Fourier features~\citep{Song2021Score} followed by a linear layer and a SiLU nonlinearity.
These are fused with the keypoint embeddings and concatenated with the descriptor $z$ inside the neural network.
The network maps the inputs to a $12$-D output space, $9$-D for a raw estimate of the rotation matrix and the remaining $3$-D for the translation vector.
The estimated rotation matrix is projected onto the nearest element of $\SO{3}$ in Frobenius norm via singular value decomposition (SVD).
The average velocity~\eqref{eq:trivialized_avg_def} and flow map~\eqref{eq:flow_map} are then computed in closed form from $\hat H_1$ via~\eqref{eq:flow_map_factors}.

\emph{Data.}
Source samples are drawn as $H_0 \sim \rho_0$, with $\rho_0$ the product of the uniform (Haar) measure on $\SO{3}$ and a standard isotropic Gaussian $\cal{N}(0, I_3)$ on $\R^3$, whereas the targets $H_1$ are the valid ground-truth grasps from the expert-annotated dataset.
Object point clouds and grasp translations are re-centered on the point-cloud centroid to enforce translation invariance by design, and each object-grasp pair is augmented by a random $\SO{3}$ rotation.
The interpolant $H_t$~\eqref{eq:geodesic_interp} uses the closed-form $\SO{3}$ exponential and logarithm, where the logarithm is evaluated through the numerically stable antipodal branch (rotation angle approaching $\pi$), so interpolation is well-defined without rejection sampling.

\emph{Objective.}
We optimize the combined objective~\eqref{eq:rmf_loss} by evaluating both the flow-matching anchor~\eqref{eq:cfm_anchor} and the semigroup consistency loss~\eqref{eq:semi_loss} on every mini-batch with $\lambda_{\mathrm{cfm}} = 1$ held fixed, while $\lambda_{\mathrm{semi}}$ is annealed linearly from $0$ to $1$ over the first $1000$ epochs so the data anchor dominates early training before self-consistency is enforced.
Both objectives are implemented with an $\ell_1$ norm on each factor instead of the squared metric norm, which performed better across all of our experiments.
For stable training, the anchor scale $1/(1-t)$ is clamped at $(1-t)_{\min} = 0.1$ and the semigroup tail weight $w(r) = (1-r)/\max(1-r,\,0.1)$ down-weights $r \to 1$.
The anchor time $t$ is drawn from the mixture $0.98\,\mathrm{Beta}(1.9, 1.0) + 0.02\,\cal{U}[0,1]$, rescaled to $[10^{-2}, 1]$, which concentrates anchor supervision near the data end of the path.
The semigroup time triplet $(s, r, t)$ is drawn interval-first: the chord length $t - s$ is sampled from $\mathrm{Beta}(1.5, 1.5)$ rescaled to $[10^{-2}, 1]$, the start time is placed uniformly so the interval fits in $[0, 1]$, and the intermediate time is the midpoint $r = (s + t)/2$; the start and intermediate times are then scaled by $1 - 10^{-2}$ as a numerical guard on the $1/(1-s)$ factor of~\eqref{eq:trivialized_avg_def}.
We additionally regress a clipped $\ell_1$ signed-distance prediction from an intermediate feature output onto ground-truth SDF values for the object, which grounds the learned flow map to the object geometry.

\begin{table}[t]
  \centering\vspace{2mm}
  \small
  \setlength{\tabcolsep}{4pt}
  \resizebox{0.65\linewidth}{!}{%
  \begin{tabular}{lccc}
    \toprule
    \multirow{2}{*}{Method} & \multicolumn{2}{c}{NFE $\downarrow$} & Latency \\
    \cmidrule(lr){2-3}
    & Per step & Total & (ms) $\downarrow$ \\
    \midrule
    SE3Dif~\citep{Urain2023SE3DiffusionFields} ($T = 70$)
    & $2\times$ & $140$ & $600.1\stdv{23.8}$ \\
    BRIDG{\scriptsize{E}}R~\citep{Chen2024BRIDGER} ($T = 40$)
    & $1\times$ & $40$ & $120.6\stdv{3.5}$ \\
    EGF~\citep{Lim2024EquiGraspFlow} ($T = 20$)
    & $4\times$ & $80$ & $187.9\stdv{1.8}$ \\
    VSIGD~\citep{Bukhari2025VariationalShapeInference} ($T = 70$)
    & $2\times$ & $140$ & $1124.4\stdv{1.4}$ \\
    \midrule
    \textbf{GraspMF} ($T = 5$)
    & $1\times$ & $\underline{5}$ & $\underline{15.5}\stdv{0.4}$ \\
    \textbf{GraspMF} ($T = 1$)
    & $1\times$ & $\mathbf{1}$ & $\mathbf{6.3}\stdv{0.2}$ \\
    \bottomrule
  \end{tabular}
  }%
  \vspace{1mm}
  \caption{Inference cost of grasp generation: network function evaluations (NFEs) and average wall-clock latency for a batch of 100 grasps per object measured on a single NVIDIA RTX 5080 GPU.
  The methods and sampling budgets $T$ follow the rows of Table~\ref{tab:main}.
  \textbf{Best} and \underline{second-best} entries are highlighted.}\label{tab:latency_nfe}
\end{table}

\emph{Optimization.}
To train the neural network, we use Adam~\citep{Kingma2015Adam} with a constant learning rate of $1 \times 10^{-4}$.
We use a mini-batch size of $2$ objects, where $200$ grasp poses are sampled per object, resulting in an effective batch size of $400$.
We train our method for $\sim 48$ hours ($\sim 10$\,K epochs) on a single NVIDIA RTX 5080 16GB GPU.

\emph{Inference.}
Grasps are generated by iterating~\eqref{eq:inference_iter} over a uniform partition of $[0, 1]$, evaluated in the factored form~\eqref{eq:flow_map_factors}, with the number of steps equal to the sampling budget $T$ of Sec.~\ref{sec:exp_budget}.
Wall-clock measurements use $N = 1024$ input points and $100$ grasps sampled per object on a single NVIDIA RTX 5080 GPU, after a $10$-iteration warm-up.

We demonstrate the core working principle of the MeanFlow formulation via an $\SO{3}$ dataset visualized as samples on the unit sphere in the form of a baseball seam in Fig.~\ref{fig:ball-seam}.
We use a lightweight MLP comprising 6 hidden layers, each of width 32, with SiLU activations.
The time arguments are concatenated at each layer, and the network head outputs a Lie algebra vector for the Flow Matching formulation, whereas the MeanFlow formulation outputs the endpoint rotation matrix estimate projected to $\SO{3}$ via SVD.
MeanFlow concentrates samples on the seam within three network evaluations, whereas Flow Matching requires tens of steps with the same backbone and training data.

\section{Experiments \& Results}\label{sec:result}

We evaluate our approach against state-of-the-art multimodal grasp generation methods in simulation and on a physical robotic grasping scenario.
In this section, we detail the experimental setup, the quantitative and qualitative results, and the ablation studies.

\begin{figure*}[t]
    \centering
    \captionsetup[subfigure]{justification=centering}
    \begin{subfigure}{\grenderwidth}
        \caption{SE3Dif~\cite{Urain2023SE3DiffusionFields}\\($T = 70$)}
    \end{subfigure}
    \hfill
    \begin{subfigure}{\grenderwidth}
        \caption{BRIDG{\scriptsize{E}}R~\cite{Chen2024BRIDGER}\\($T = 40$)}
    \end{subfigure}
    \hfill
    \begin{subfigure}{\grenderwidth}
        \caption{EGF~\cite{Lim2024EquiGraspFlow}\\($T = 20$)}
    \end{subfigure}
    \hfill
    \begin{subfigure}{\grenderwidth}
        \caption{VSIGD~\cite{Bukhari2025VariationalShapeInference}\\($T = 70$)}
    \end{subfigure}
    \hfill
    \begin{subfigure}{\grenderwidth}
        \caption{\textbf{GraspMF}\\($T = 5$)}
    \end{subfigure}
    \hfill
    \begin{subfigure}{\grenderwidth}
        \caption{\textbf{GraspMF}\\($T = 1$)}
    \end{subfigure}
    \newline
    \begin{subfigure}{\grenderwidth}
        \begin{overpic}[width=\textwidth,tics=10]{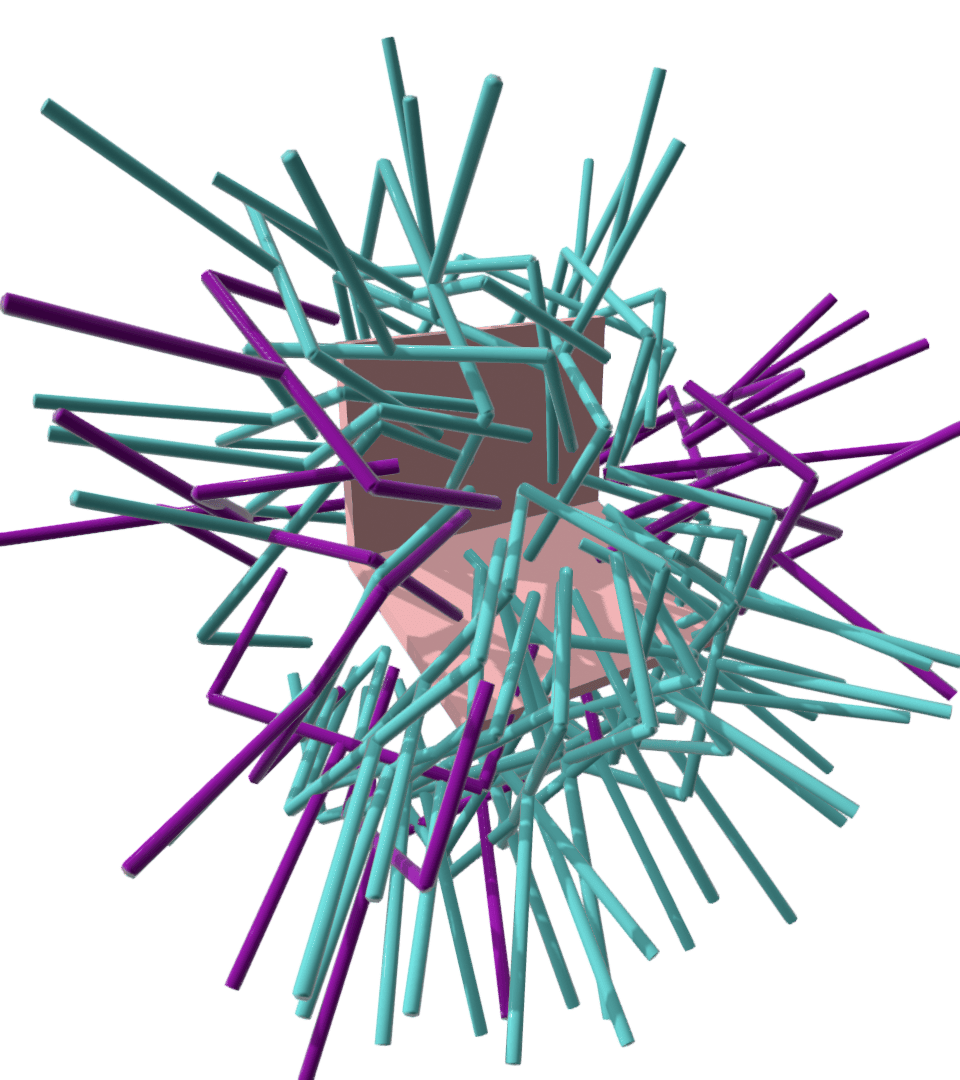}
        \end{overpic}
    \end{subfigure}
    \hfill
    \begin{subfigure}{\grenderwidth}
        \begin{overpic}[width=\textwidth,tics=10]{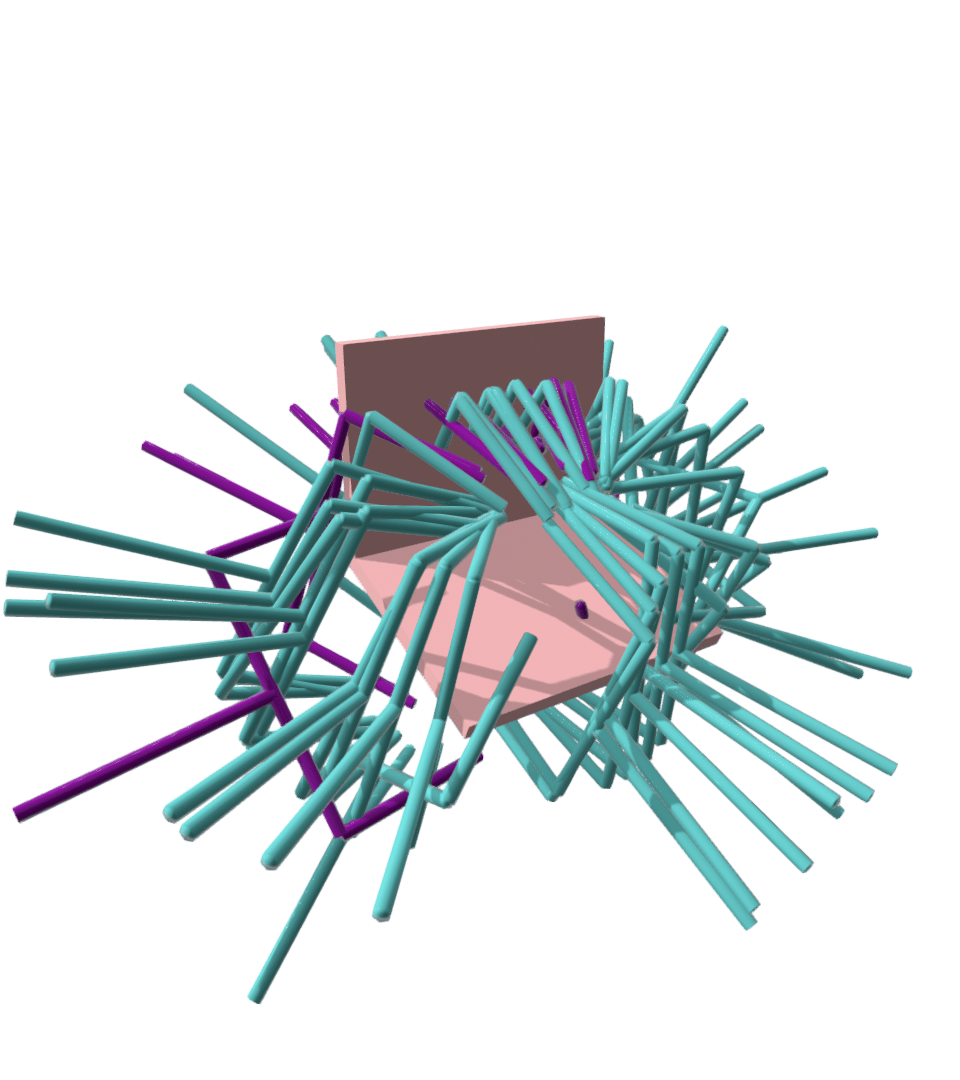}
        \end{overpic}
    \end{subfigure}
    \hfill
    \begin{subfigure}{\grenderwidth}
        \begin{overpic}[width=\textwidth,tics=10]{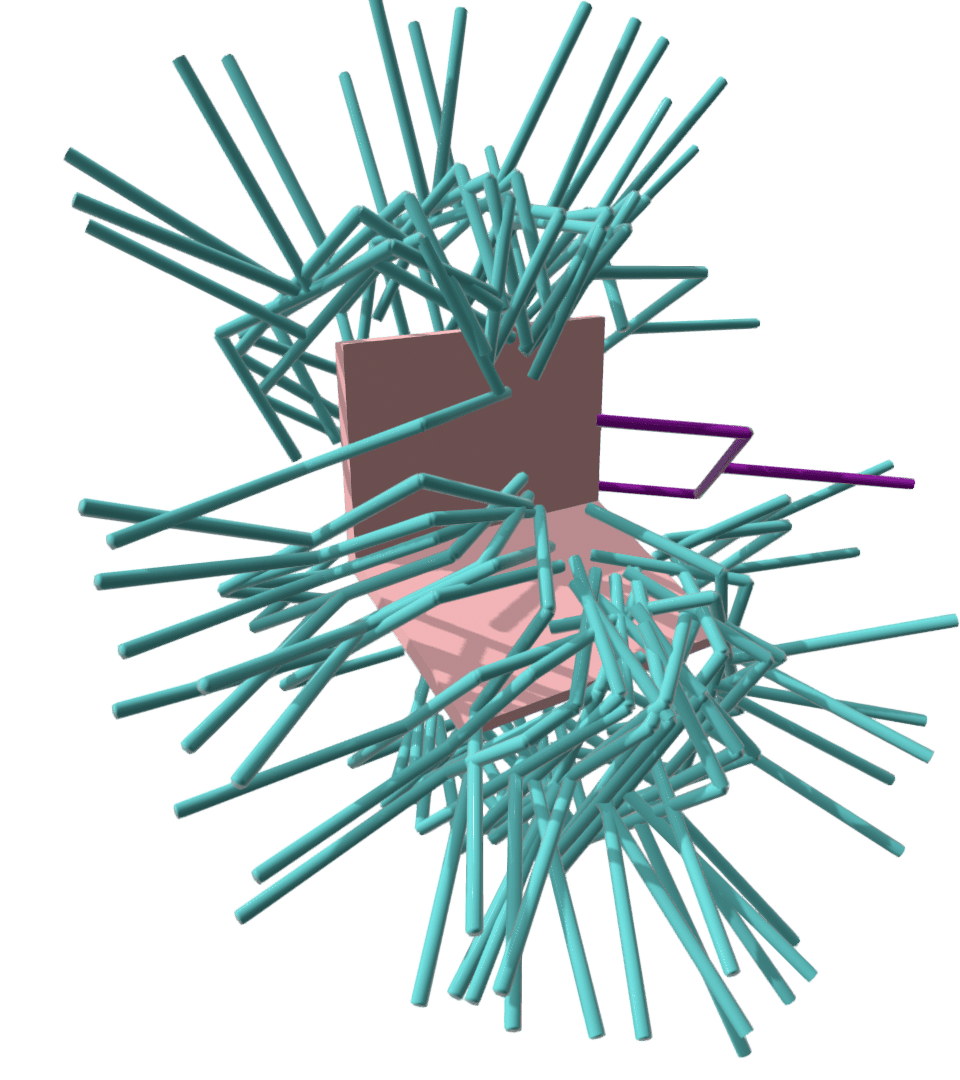}
        \end{overpic}
    \end{subfigure}
    \hfill
    \begin{subfigure}{\grenderwidth}
        \begin{overpic}[width=\textwidth,tics=10]{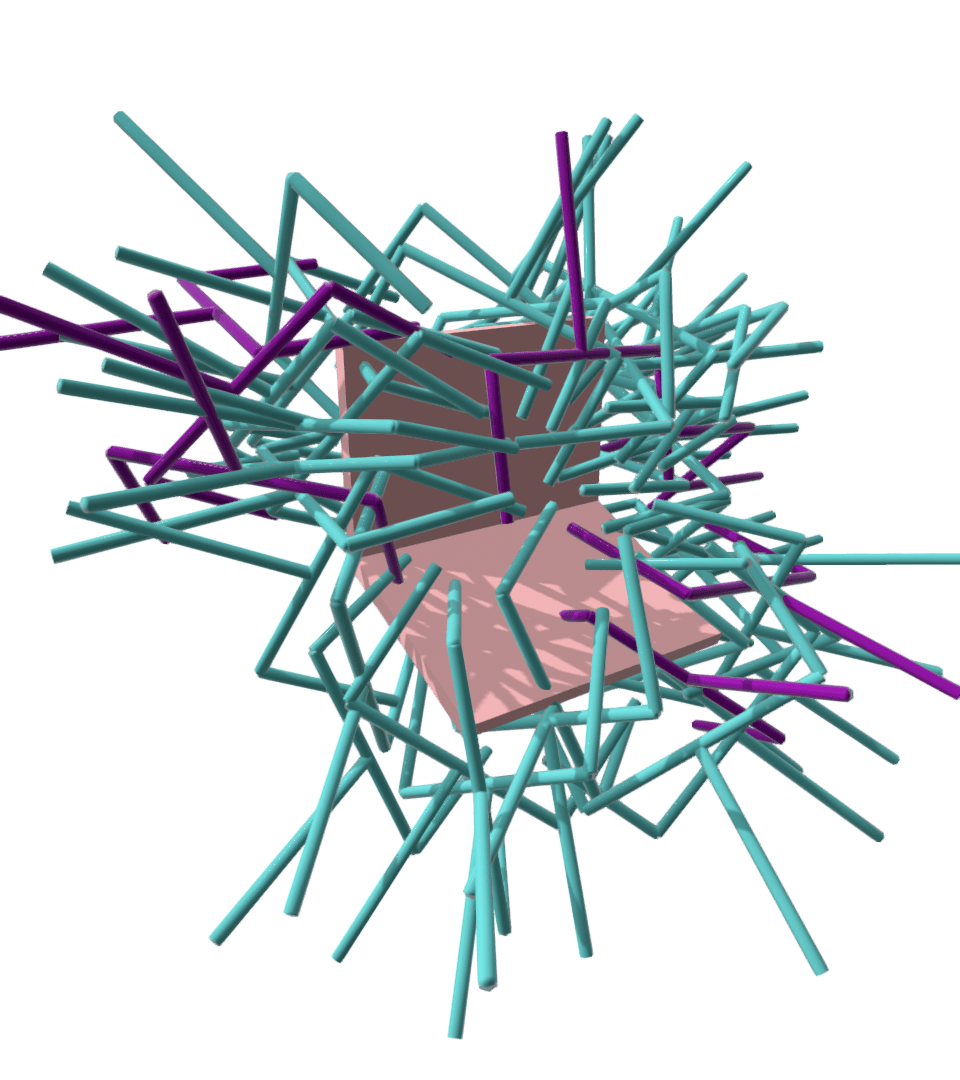}
        \end{overpic}
    \end{subfigure}
    \hfill
    \begin{subfigure}{\grenderwidth}
        \begin{overpic}[width=\textwidth,tics=10]{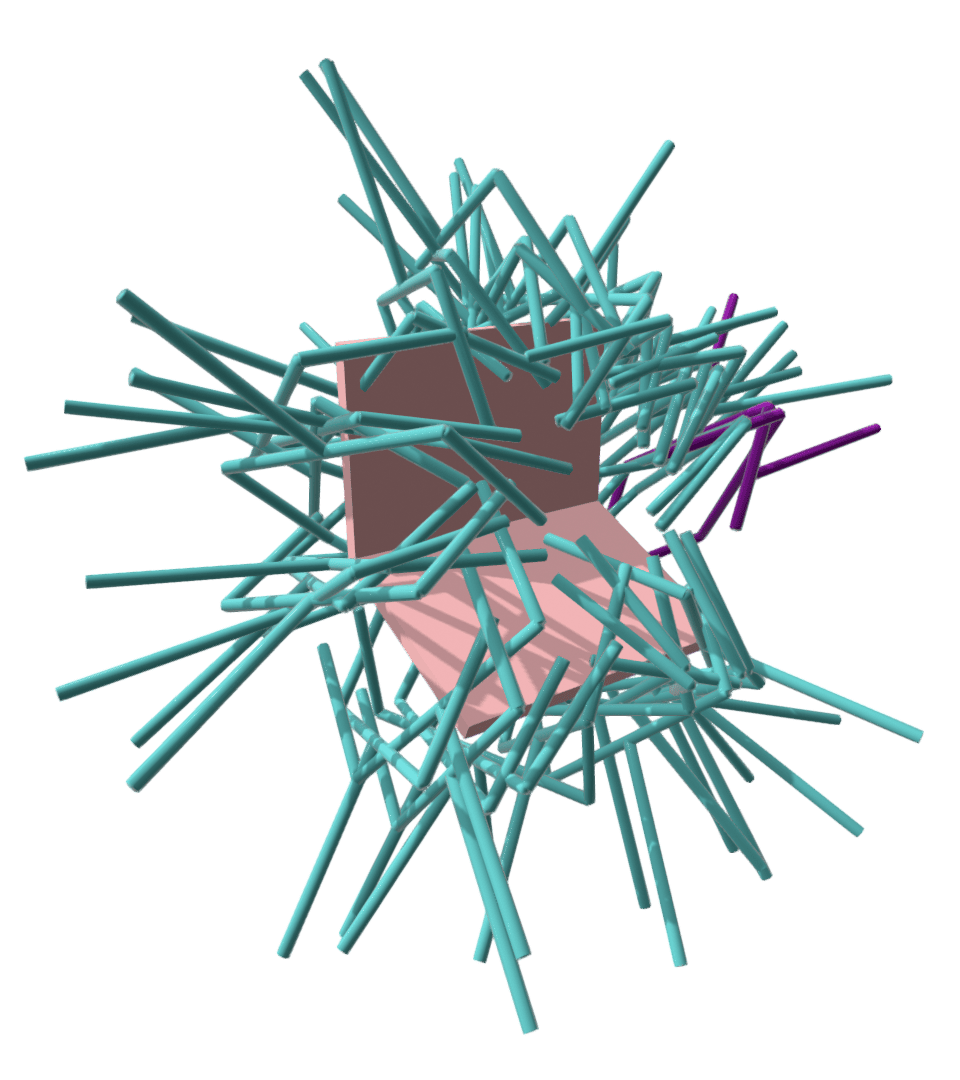}
        \end{overpic}
    \end{subfigure}
    \hfill
    \begin{subfigure}{\grenderwidth}
        \begin{overpic}[width=\textwidth,tics=10]{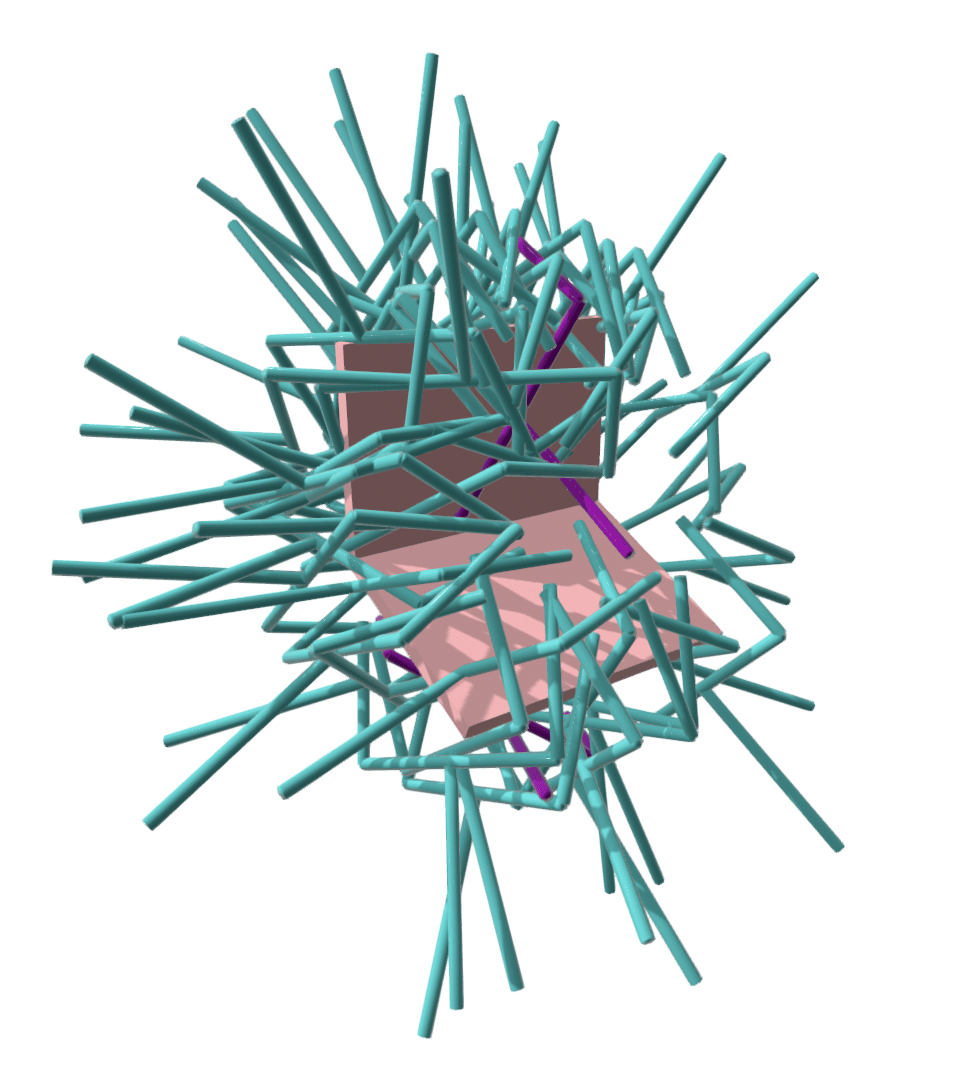}
        \end{overpic}
    \end{subfigure}
    \newline
    \begin{subfigure}{\grenderwidth}
        \begin{overpic}[width=\textwidth,tics=10]{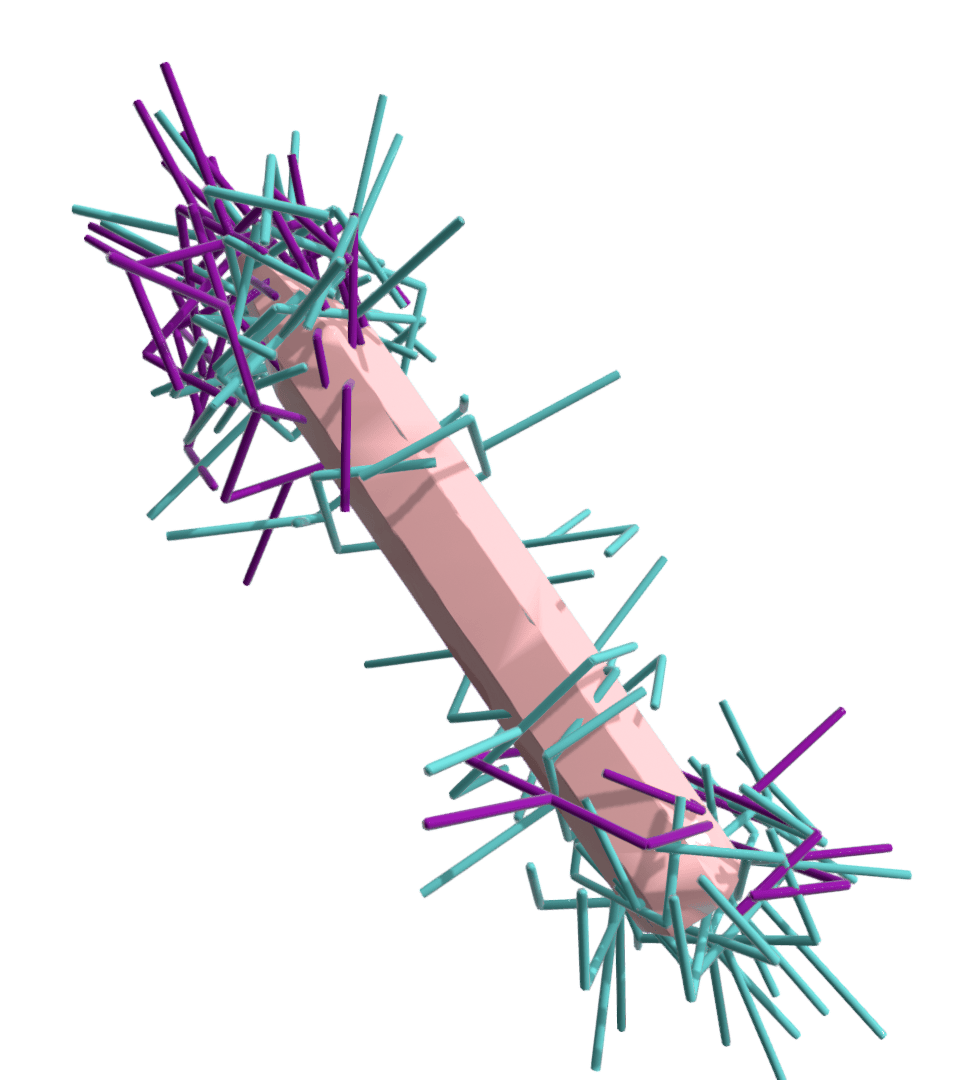}
        \end{overpic}
    \end{subfigure}
    \hfill
    \begin{subfigure}{\grenderwidth}
        \begin{overpic}[width=\textwidth,tics=10]{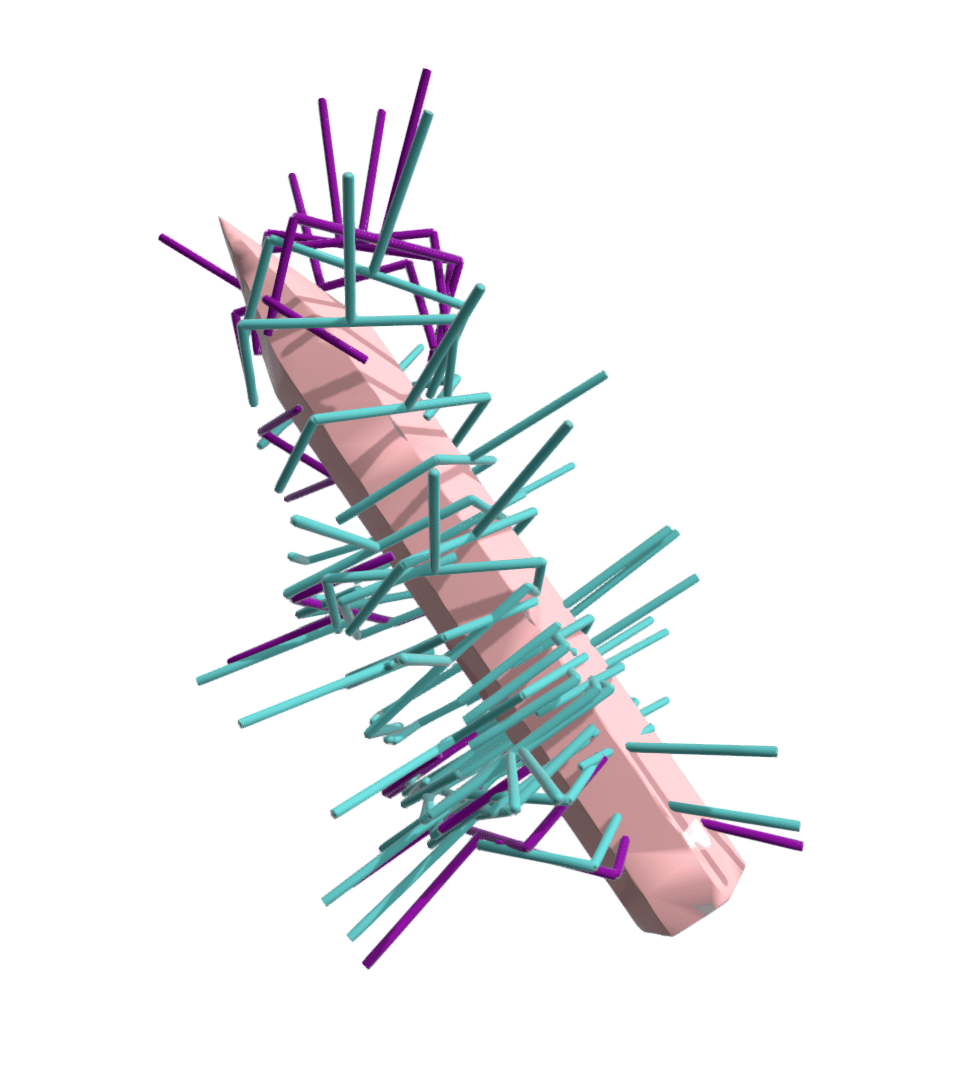}
        \end{overpic}
    \end{subfigure}
    \hfill
    \begin{subfigure}{\grenderwidth}
        \begin{overpic}[width=\textwidth,tics=10]{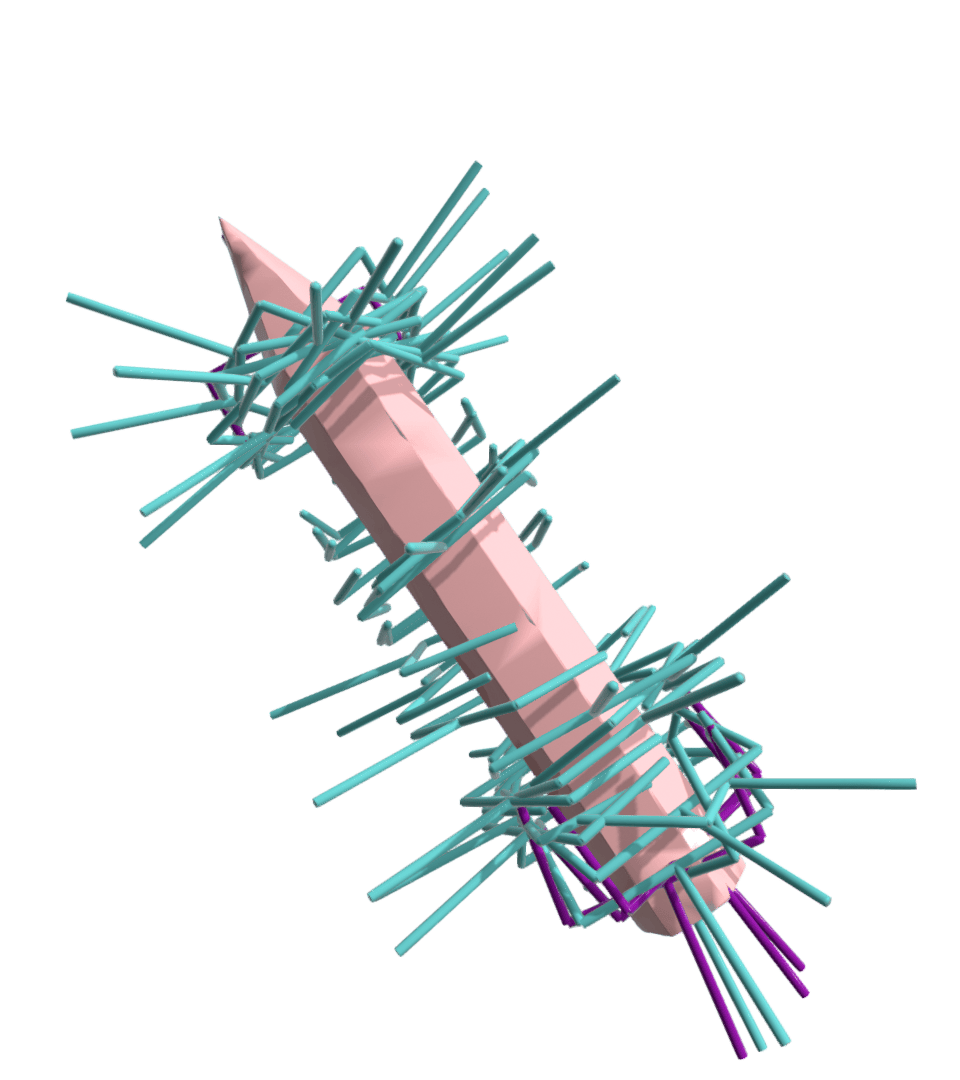}
        \end{overpic}
    \end{subfigure}
    \hfill
    \begin{subfigure}{\grenderwidth}
        \begin{overpic}[width=\textwidth,tics=10]{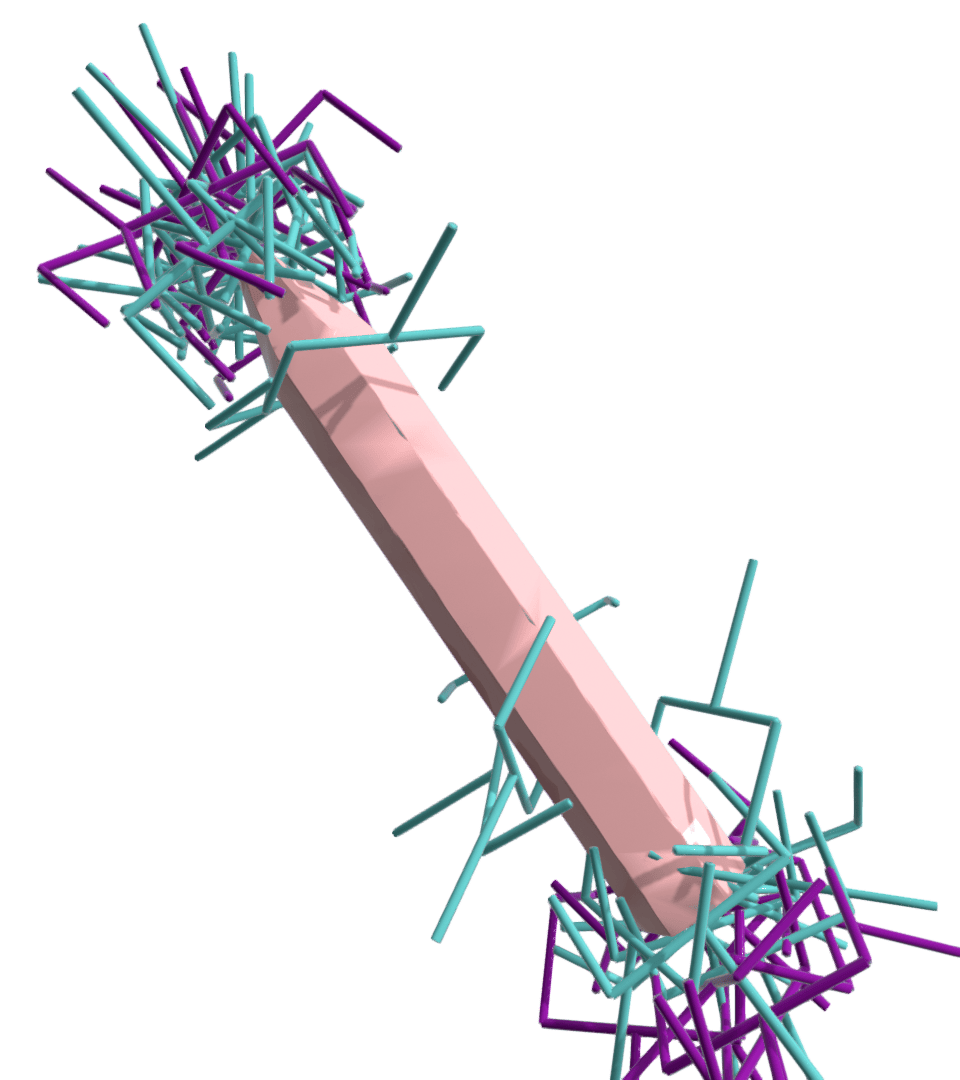}
        \end{overpic}
    \end{subfigure}
    \hfill
    \begin{subfigure}{\grenderwidth}
        \begin{overpic}[width=\textwidth,tics=10]{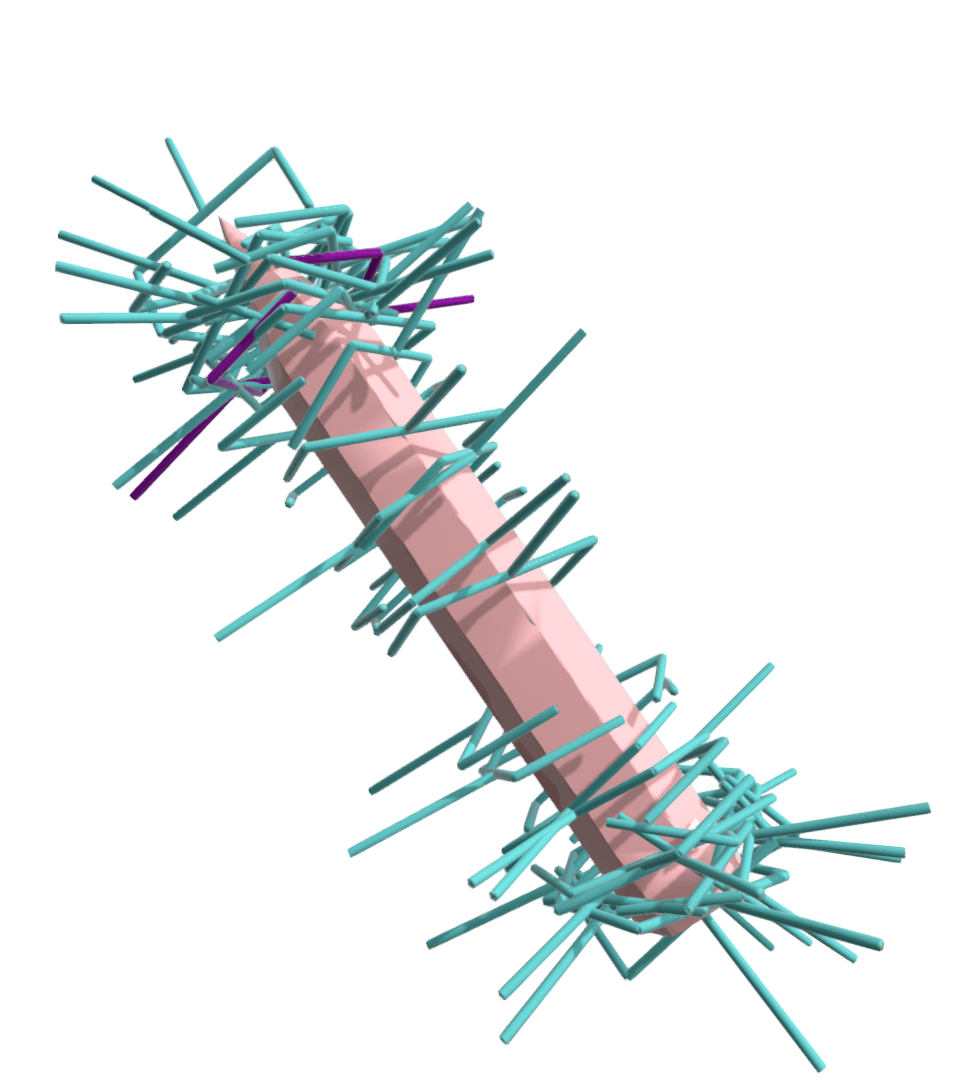}
        \end{overpic}
    \end{subfigure}
    \hfill
    \begin{subfigure}{\grenderwidth}
        \begin{overpic}[width=\textwidth,tics=10]{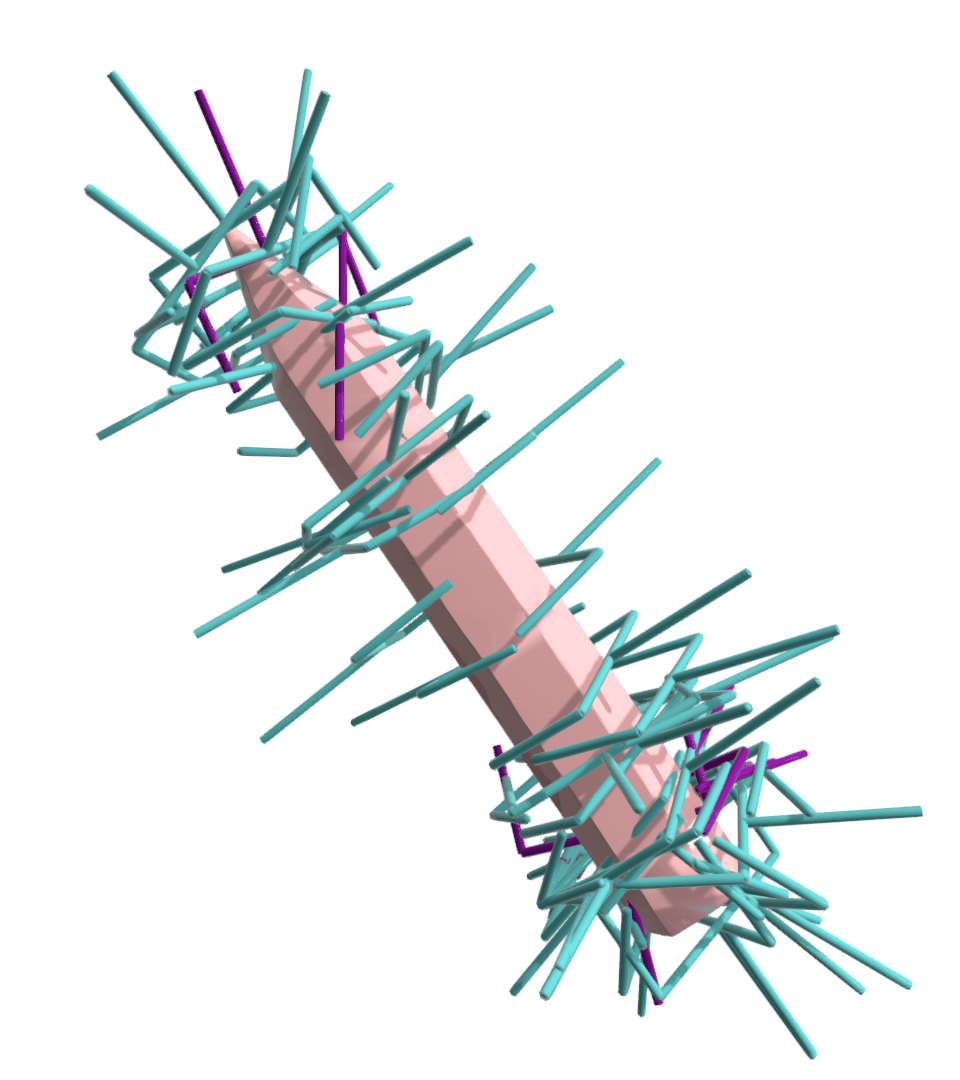}
        \end{overpic}
    \end{subfigure}
    \newline
    \begin{subfigure}{\grenderwidth}
        \begin{overpic}[width=\textwidth,tics=10]{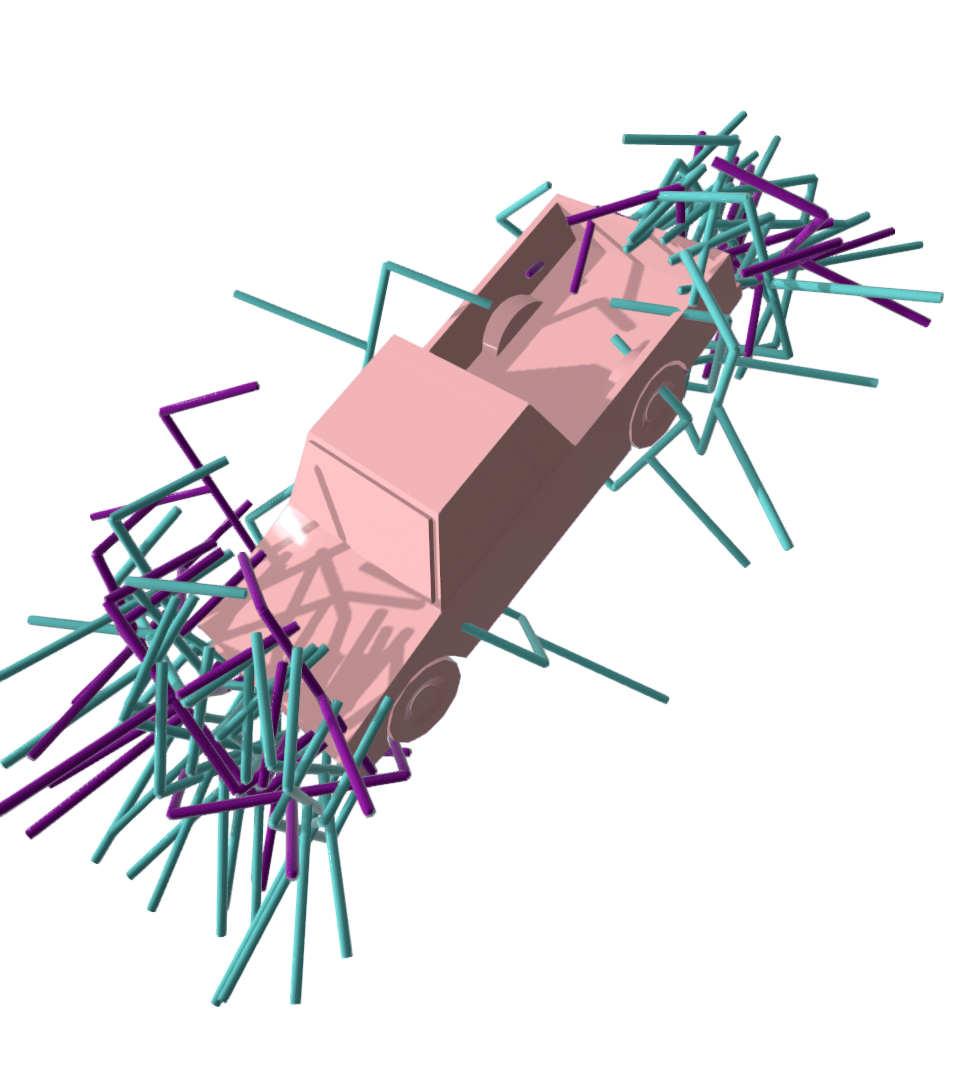}
        \end{overpic}
    \end{subfigure}
    \hfill
    \begin{subfigure}{\grenderwidth}
        \begin{overpic}[width=\textwidth,tics=10]{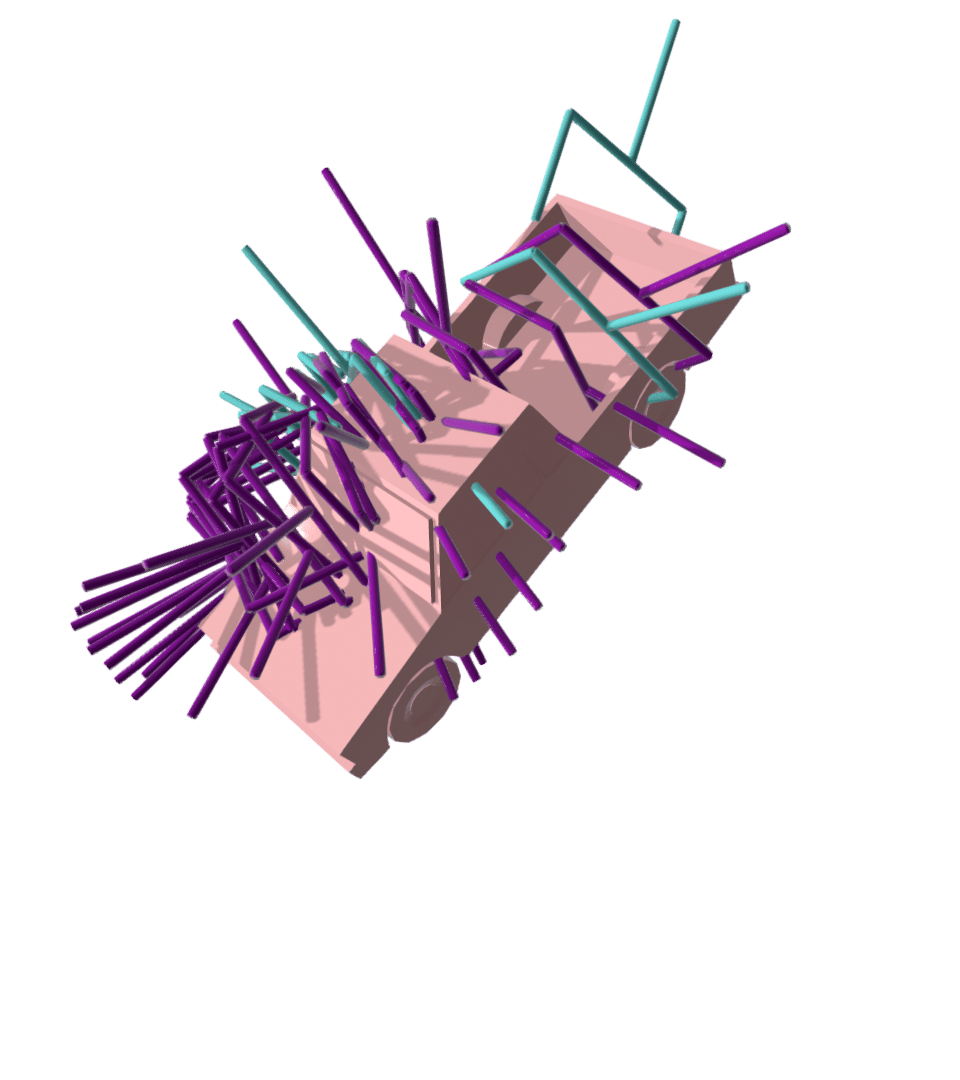}
        \end{overpic}
    \end{subfigure}
    \hfill
    \begin{subfigure}{\grenderwidth}
        \begin{overpic}[width=\textwidth,tics=10]{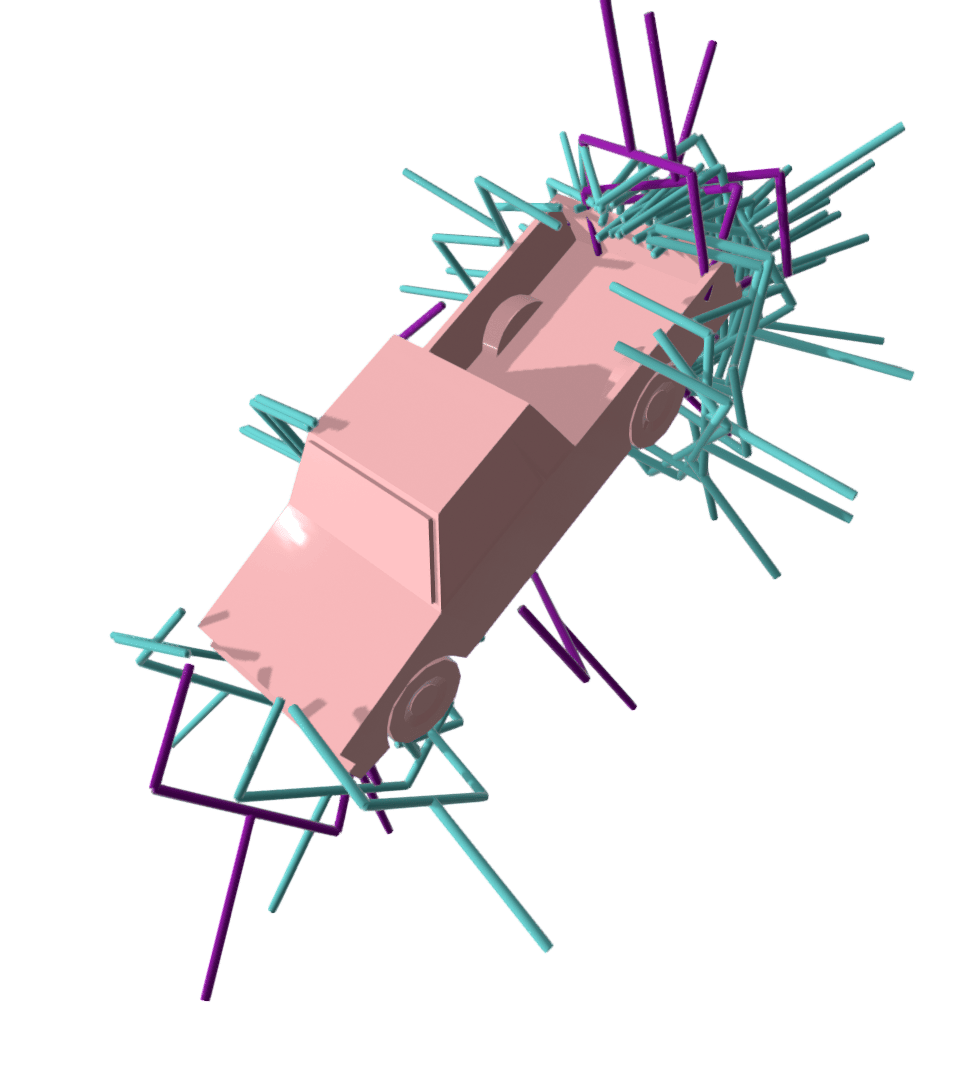}
        \end{overpic}
    \end{subfigure}
    \hfill
    \begin{subfigure}{\grenderwidth}
        \begin{overpic}[width=\textwidth,tics=10]{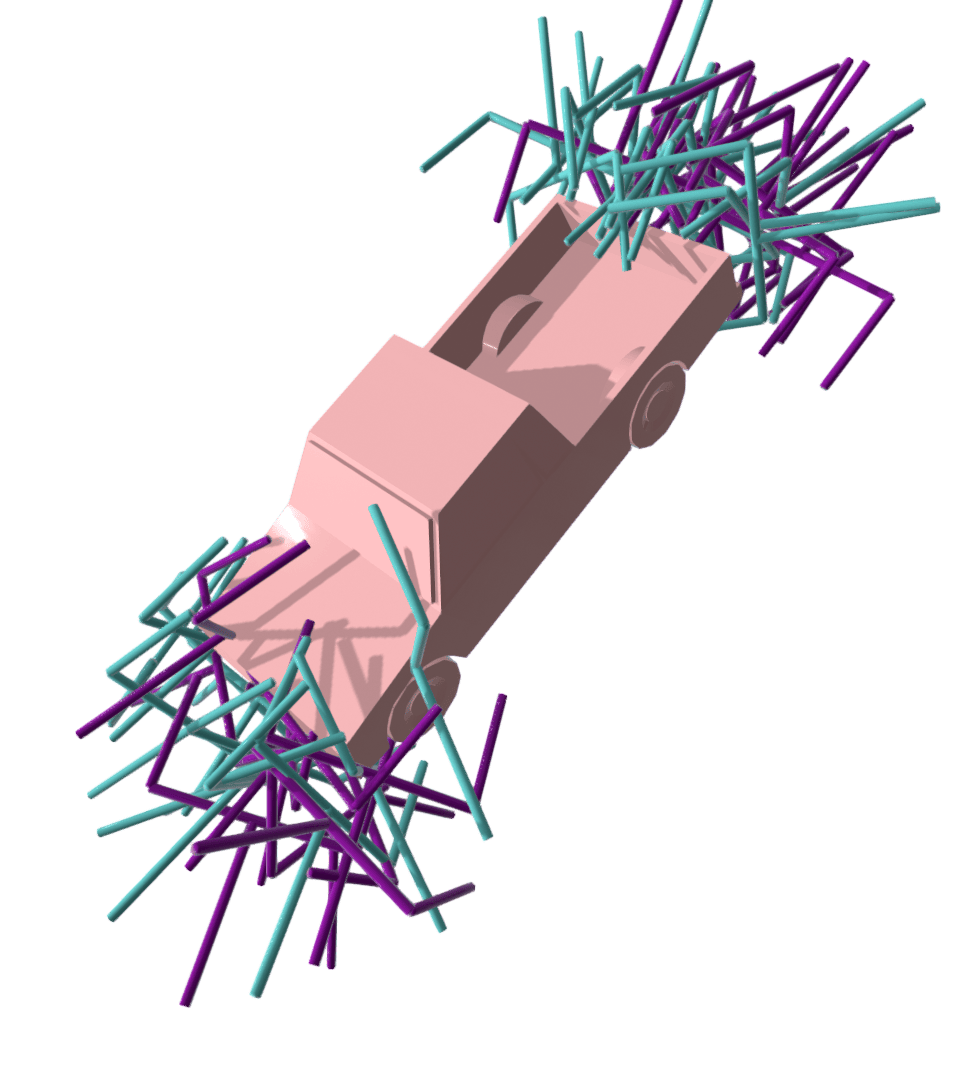}
        \end{overpic}
    \end{subfigure}
    \hfill
    \begin{subfigure}{\grenderwidth}
        \begin{overpic}[width=\textwidth,tics=10]{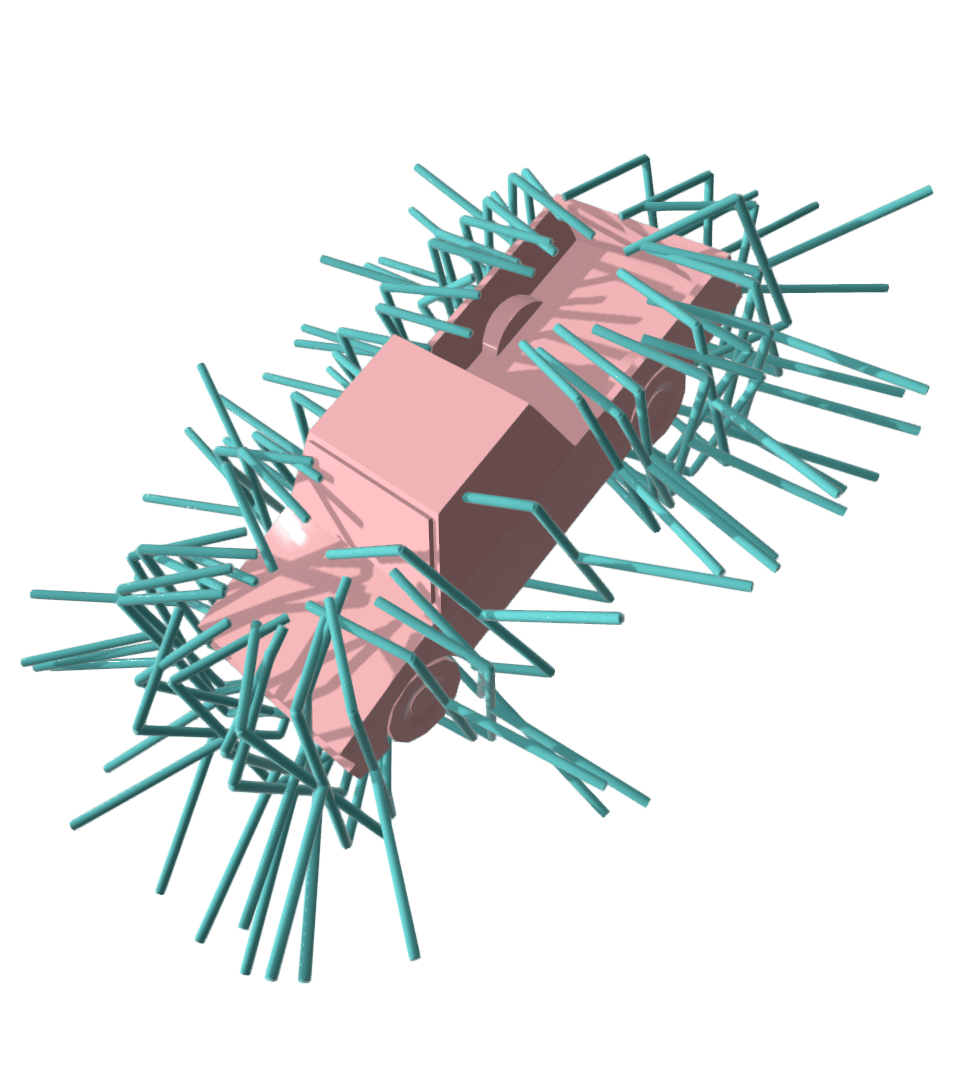}
        \end{overpic}
    \end{subfigure}
    \hfill
    \begin{subfigure}{\grenderwidth}
        \begin{overpic}[width=\textwidth,tics=10]{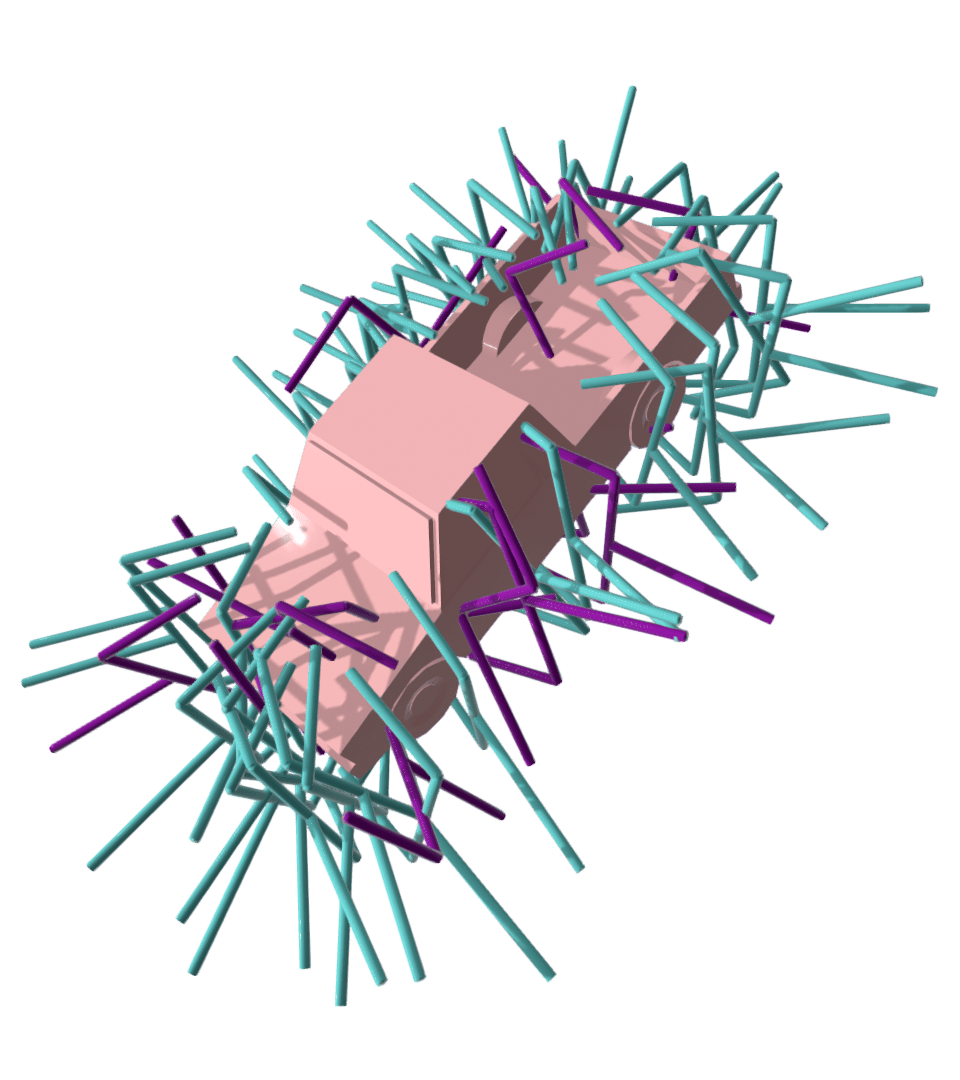}
        \end{overpic}
    \end{subfigure}
    \caption{\label{fig:grasp-renders}
    Qualitative comparison of generated grasps for Laptop, Pencil, and Car (in order).
    \textcolor{grasp1}{\bf Dark cyan} indicates successful grasps while \textcolor{grasp0}{\bf dark purple} indicates failures.
    The columns follow the rows of Table~\ref{tab:main}: each comparison method is shown at the sampling budget of its original formulation, and GraspMF at its low-step ($T = 5$) and single-step ($T = 1$) settings.
    GraspMF generates more successful and stable grasps with broader grasp-mode coverage across diverse object geometries, as evaluated in IsaacGym.
    }\vspace{-2mm}
\end{figure*}

\subsection{Experimental Setup}\label{ssec:exp_setup}

For evaluating grasp generation performance, we build on the experimentation and evaluation protocol of prior work on deep generative grasping~\citep{Urain2023SE3DiffusionFields, Lim2024EquiGraspFlow, Chen2024BRIDGER, Bukhari2025VariationalShapeInference}.
Here, we provide the details of our experimental setup.

\subsubsection{Dataset}\label{sssec:dataset}

For grasp training and evaluation, we use the object meshes and their expert grasp annotations available in the ACRONYM dataset~\citep{Eppner2021ACRONYM}.
We consider the following object shape categories from the dataset: \texttt{Book}, \texttt{Bottle}, \texttt{Bowl}, \texttt{Cap}, \texttt{CellPhone}, \texttt{Cup}, \texttt{Hammer}, \texttt{Mug}, \texttt{Scissors}, and \texttt{Shampoo}.
This amounts to $416$ object instances, with $\sim 780$\,K valid grasps in total.
From this data subset, 90\% of the instances from each category are used for training all methods and the remaining 10\% are reserved for evaluating in-domain (ID) performance.
We additionally consider a disjoint set of object categories from the ACRONYM dataset for evaluating out-of-domain (OOD) generalization: \texttt{Car}, \texttt{Donut}, \texttt{Laptop}, \texttt{Pencil}, \texttt{RubiksCube}, and \texttt{Spoon}.
This split isolates in-domain from out-of-domain generalization while retaining sufficient training data for learning grasp generation.

\subsubsection{Baseline Methods}\label{sssec:baselines}

For evaluating the efficacy of GraspMF, we compare against the following state-of-the-art grasp generation methods that consider learning a distribution over $\SE{3}$ grasp poses:
\begin{itemize}
    \item $\SE{3}$-DiffusionFields~\citep{Urain2023SE3DiffusionFields} (SE3Dif), which learns a grasp energy field on $\SE{3}$ via denoising score matching. This method serves as the diffusion-based baseline in our evaluation.
    \item EquiGraspFlow~\citep{Lim2024EquiGraspFlow} (EGF), which learns a Lie-algebra-valued vector field on $\SE{3}$ via flow matching, with a neural network architecture that promotes $\SE{3}$ equivariance. This method is the flow-based baseline in our evaluation.
    \item BRIDG{\scriptsize{E}}R~\citep{Chen2024BRIDGER}, which initializes samples from an informative prior and transports them to the target distribution via a linear interpolant function, providing an approach for fast grasp generation.
    \item VSIGD~\citep{Bukhari2025VariationalShapeInference}, which augments the score-based generative modeling of SE3Dif with shape-space pretraining and autoencoding, for robust grasp generation under noisy observations, providing a reference point for high OOD performance.
\end{itemize}
For a meaningful comparison, all approaches are configured at the sampling budgets reported by their authors for noise-to-data transport.
We evaluate the impact of the sampling budget on grasping performance in Sec.~\ref{sec:exp_budget}.

\begin{figure*}[t]
  \begin{subfigure}{\textwidth}
    \centering
    \includegraphics[width=0.4\linewidth]{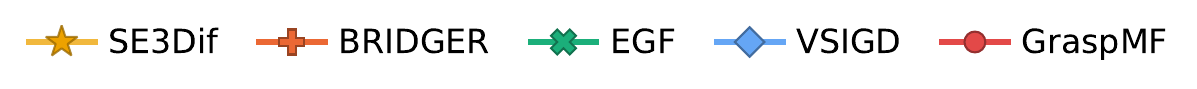}\label{fig:legend}
  \end{subfigure}
  \begin{subfigure}{0.24\textwidth}
    \centering
    \includegraphics[width=\linewidth]{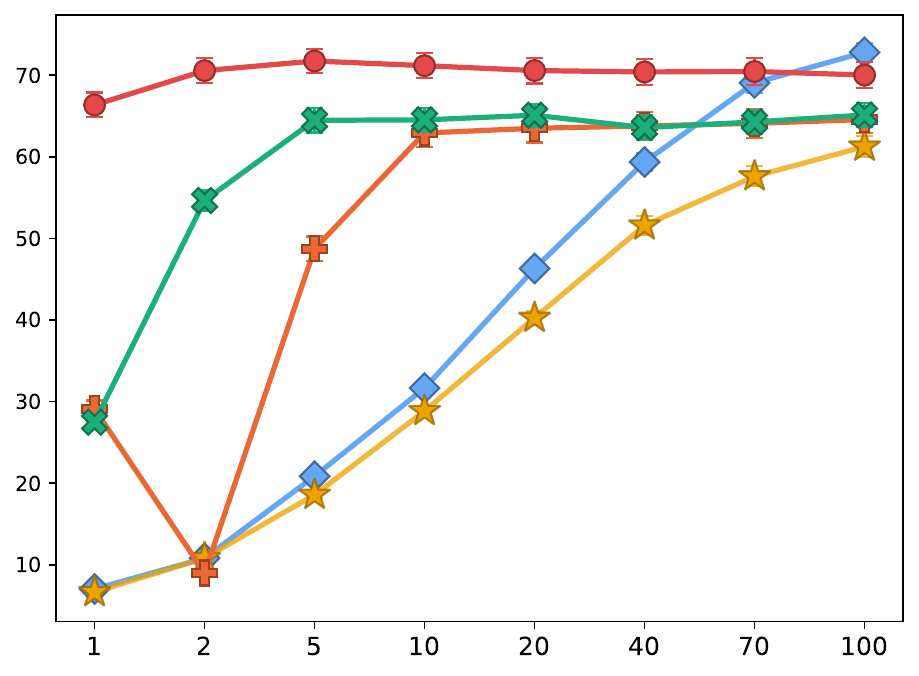}
    \caption{SR $(\%)$ $\uparrow$}\label{fig:budget_sr}
  \end{subfigure}
  \hfill
  \begin{subfigure}{0.24\textwidth}
    \centering
    \includegraphics[width=\linewidth]{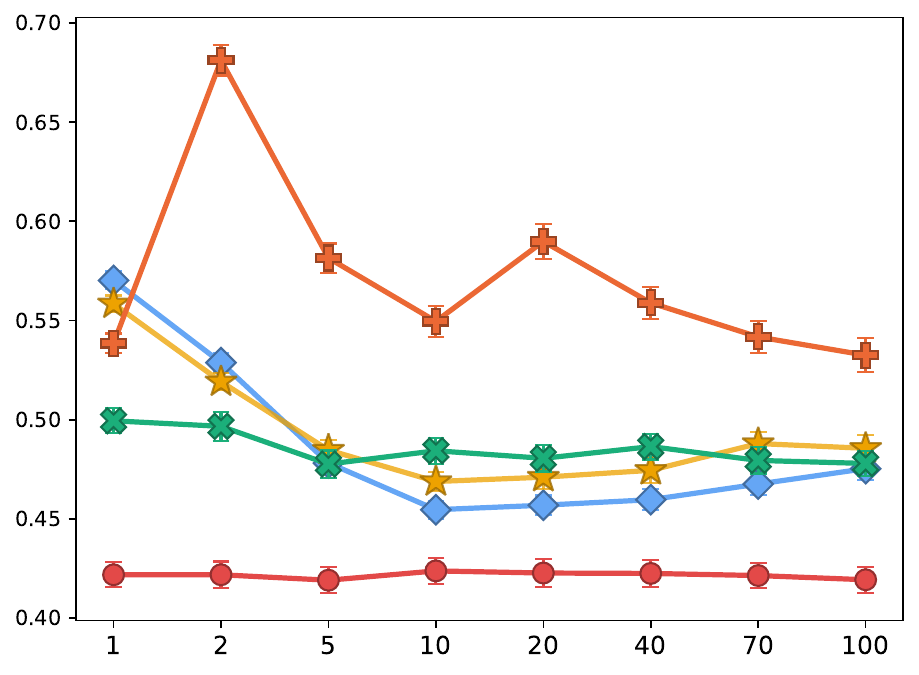}
    \caption{EMD $\downarrow$}\label{fig:budget_emd}
  \end{subfigure}
  \hfill
  \begin{subfigure}{0.24\textwidth}
    \centering
    \includegraphics[width=\linewidth]{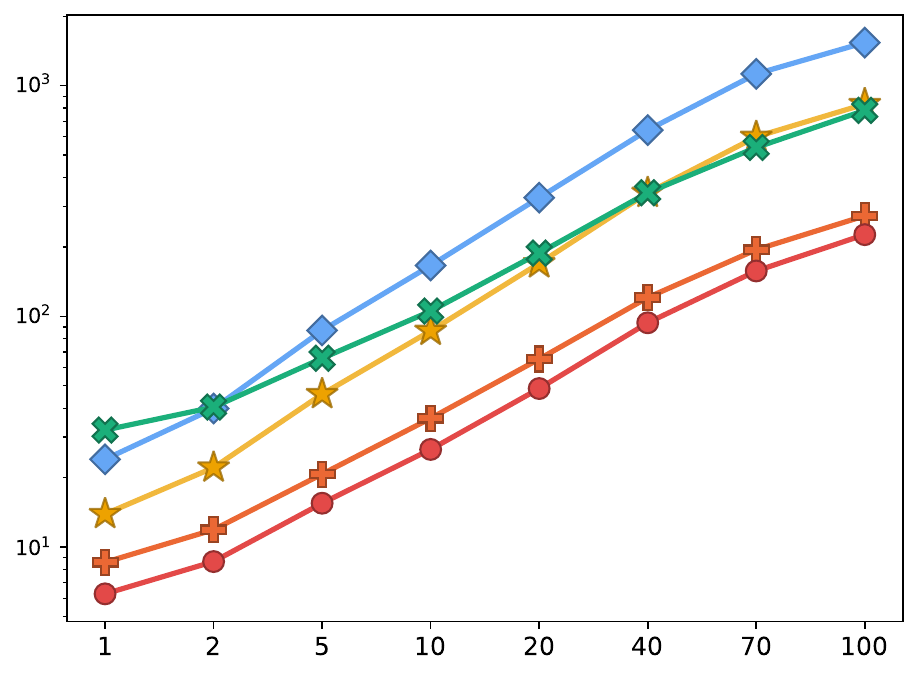}
    \caption{Latency (ms, log scale) $\downarrow$}\label{fig:budget_time}
  \end{subfigure}
  \hfill
  \begin{subfigure}{0.24\textwidth}
    \centering
    \includegraphics[width=\linewidth]{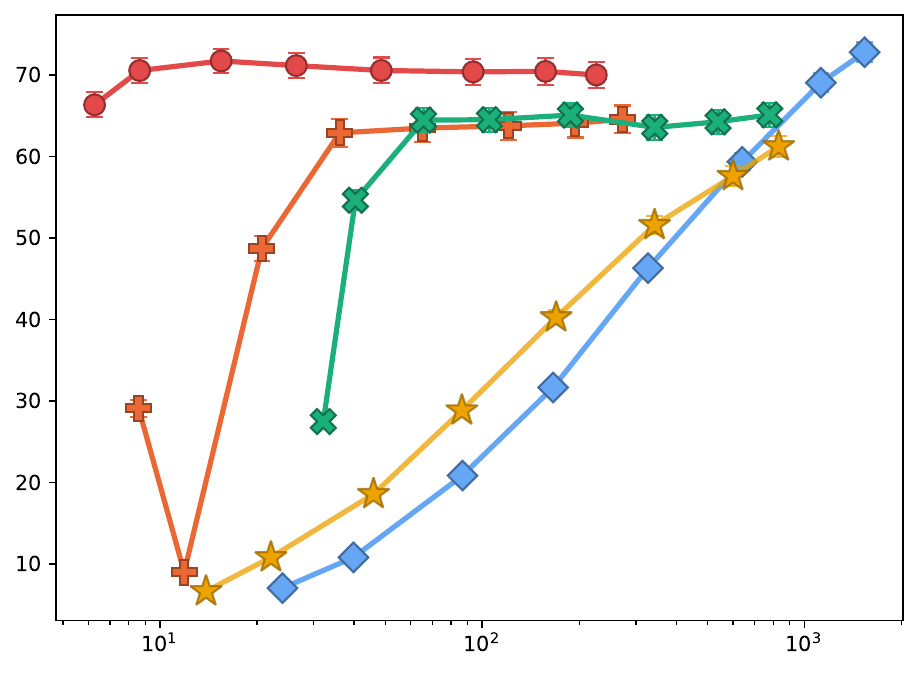}
    \caption{SR vs Latency}\label{fig:sr_vs_time}
  \end{subfigure}
  \caption{Grasping performance on the OOD evaluation protocol across the sampling budget $T \in \{1, 2, 5, 10, 20, 40, 70, 100\}$.
  SR in \% (\ref{fig:budget_sr}), EMD (\ref{fig:budget_emd}), and inference latency in $\log$ scale (\ref{fig:budget_time}) are plotted against sampling budgets, and SR is plotted against inference latency in $\log$ scale (\ref{fig:sr_vs_time}) where the top-left corner approaches the Pareto frontier.}\label{fig:budget}
\end{figure*}

\subsubsection{Performance Metrics}\label{sssec:metrics}

We evaluate all methods on three performance metrics, computed over 100 sampled grasps per object instance:
\begin{itemize}
  \item \emph{Success rate} (SR): Sampled grasps are executed in Isaac Gym~\citep{Makoviychuk2021IsaacGym} via the open-grip $\rightarrow$ approach  $\rightarrow$ close-grip  $\rightarrow$ lift protocol of~\citet{Eppner2021ACRONYM}, and counted as a success if the object remains in the gripper after a $5$\,s shake.
  This metric quantifies the effective grasping performance.
  \item \emph{Earth Mover's Distance (EMD)}: This metric is computed empirically between the 100 sampled grasps and 100 ground-truth samples using the pose distance $d((R, p), (R', p')) = \|\log(R^\top R')^\vee\|_2 + \lambda\|p - p'\|_2$, which combines the $\SO{3}$ geodesic distance with a weighted Euclidean translation term; the weight $\lambda = 0.1$ equates $1$\,m of translation with $5.7^\circ$ of rotation.
  This metric serves as a proxy for distribution coverage.
  \item \emph{Inference latency}: We also report the averaged wall-clock time in milliseconds (ms) to generate a batch of 100 grasps per object in parallel, on a single NVIDIA RTX 5080 GPU.
\end{itemize}

\subsection{Benchmarking Grasp Generation}\label{sec:exp_quant}

We evaluate the grasp generation performance of GraspMF against baseline methods on objects from the ACRONYM dataset, under the in-domain (ID) and out-of-domain (OOD) evaluation protocols described in Sec.~\ref{sssec:dataset}.
Table~\ref{tab:main} reports SR in simulation and empirical EMD to ground-truth grasp poses, and Table~\ref{tab:latency_nfe} reports the corresponding NFE counts and sampling latency.
For GraspMF, we provide performance for one-step ($T = 1$) and few-step ($T = 5$) sampling modes.
For both ID and OOD evaluation, GraspMF with few-step generation achieves the highest SR (ID: 87.40\%, OOD: 71.73\%), while attaining the best EMD for OOD (0.4191) and competitive EMD for ID (0.3702), at a $15.5$\,ms inference latency.
The closest baselines are EGF for ID and VSIGD for OOD, with $1.9$--$2.7$ percentage points lower SR and $0.02$--$0.05$ higher EMD.
Furthermore, EGF with a higher sampling budget and an equivariance-focused network architecture incurs a $\sim 12\times$ higher latency at $187.9$\,ms, while VSIGD with its shape inference design incurs a $\sim 73\times$ higher latency at $1124.4$\,ms (Table~\ref{tab:latency_nfe}).
By contrast, GraspMF uses the efficient network design of SE3Dif as the backbone while attaining higher grasping performance at a fraction of the inference costs due to the MeanFlow formulation.
Furthermore, GraspMF with single-step generation attains a competitive SR (ID: 81.11\%, OOD: 66.34\%) and EMD (ID: 0.3698, OOD: 0.4218), at a $6.3$\,ms inference latency.
SE3Dif and VSIGD incur the highest inference costs because their energy-based formulation of denoising score matching requires 2 NFEs per time step in a predictor--corrector sampling scheme, whereas EGF requires 4 NFEs per step via a fourth-order Runge-Kutta MK integration scheme~\cite{Munthe1998Runge}.
At their reported sampling budgets, these three methods expend 140, 140, and 80 NFEs respectively, against 5 for GraspMF at $T = 5$ and 1 at $T = 1$.
GraspMF therefore attains high SR and competitive EMD, while its low NFE count and efficient network design enable millisecond-scale inference.

Fig.~\ref{fig:grasp-renders} compares the grasps generated for representative objects from the following classes of the OOD split: \texttt{Laptop}, \texttt{Pencil}, and \texttt{Car}.
GraspMF produces a larger fraction of successful grasps (dark cyan) compared to the baseline methods, and the successful grasps span several distinct approach directions and contact regions on each object.
Hence, the sampler attains broader coverage of multiple grasp modes rather than collapsing onto a single pose, mirroring the performance metrics of Table~\ref{tab:main}.
The high success rate and mode coverage persist in the single-step mode (Fig.~\ref{fig:grasp-renders}, Column~f), indicating that one network evaluation can achieve competitive end performance, and that its $6.3$\,ms latency admits closed-loop replanning that may achieve better performance.

\subsection{Performance Across Sampling Budgets}\label{sec:exp_budget}

Besides the absolute performance metrics of Table~\ref{tab:main}, we also quantify the impact of sampling budgets and the extent of performance degradation when operating in the few-step regime.
Fig.~\ref{fig:budget} plots SR, EMD, and inference latency as functions of the sampling budget $T \in \{1, 2, 5, 10, 20, 40, 70, 100\}$.
GraspMF is stable across sampling budgets: its OOD SR averages $70.15\%$ with a standard deviation of $1.63$.
This indicates that the semigroup consistency objective regularizes the learned flow map sufficiently to enable few-step generation.
By contrast, the performance of the baselines degrades sharply at low budgets: at $T = 1$ and $T = 5$, SR drops to $6.65\%$ and $18.58\%$ for SE3Dif, and to $7.04\%$ and $20.86\%$ for VSIGD, respectively.
SR for EGF drops to $27.48\%$ at $T = 1$ but remains consistent for $T \geq 5$, while staying strictly below the SR of GraspMF at every budget.
BRIDG{\scriptsize{E}}R behaves erratically before stabilizing for $T \geq 10$, indicating that although the method speeds up inference relative to diffusion-based methods, it is not designed for the few-step regime.
Fig.~\ref{fig:budget_time} shows that GraspMF has the lowest inference latency of all methods at every sampling budget, a consequence of its lightweight backbone and of the single network evaluation it expends per step, in contrast to the multi-evaluation steps of SE3Dif, VSIGD, and EGF (Table~\ref{tab:latency_nfe}).
We also plot SR against inference latency in Fig.~\ref{fig:sr_vs_time}, where GraspMF occupies the upper-left region of the plane, tracing the Pareto frontier for the methods under consideration.
Hence, GraspMF achieves the best trade-off between grasping performance and inference latency.

\subsection{Ablations}\label{sec:exp_ablation}

\begin{table}[t]
  \centering\vspace{2mm}
  \small
  \setlength{\tabcolsep}{3pt}
  \resizebox{\linewidth}{!}{%
  \begin{tabular}{lcccc}
    \toprule
    \multirow{2}{*}{Variant} & \multicolumn{2}{c}{SR $(\%)$ $\uparrow$} & \multicolumn{2}{c}{EMD $\downarrow$} \\
    \cmidrule(lr){2-3} \cmidrule(lr){4-5}
    & ID & OOD & ID & OOD \\
    \midrule
    \textbf{GraspMF} & $87.40\stdv{17.43}$ & $\mathbf{71.73}\stdv{25.57}$ & $\mathbf{0.3702}\stdv{0.0942}$ & $\mathbf{0.4191}\stdv{0.1108}$ \\
    GraspMF-NoSDF & $83.62\stdv{17.90}$ & $66.60\stdv{22.78}$ & $0.3865\stdv{0.1163}$ & $0.4278\stdv{0.1007}$ \\
    GraspMF-NoSched & $85.02\stdv{19.12}$ & $65.95\stdv{24.81}$ & $0.3777\stdv{0.1025}$ & $0.4235\stdv{0.1057}$ \\
    GraspMF-GS & $\mathbf{87.51}\stdv{16.08}$ & $67.05\stdv{26.62}$ & $0.3790\stdv{0.1022}$ & $0.4259\stdv{0.1122}$ \\
    GraspMF-Decomp & $62.25\stdv{20.43}$ & $52.29\stdv{19.05}$ & $0.3848\stdv{0.1084}$ & $0.4265\stdv{0.0904}$ \\
    \bottomrule
  \end{tabular}%
  }
  \vspace{1mm}
  \caption{Ablations of GraspMF at sampling budget $T = 5$.
  GraspMF-NoSDF removes the auxiliary signed-distance regression; GraspMF-NoSched holds the semigroup weight constant instead of annealing it; GraspMF-GS replaces the SVD rotation projection with symmetric Gram--Schmidt orthonormalization; GraspMF-Decomp uses the decomposed objective of~\citet{Zhong2026RMFManifolds} in place of the semigroup loss~\eqref{eq:semi_loss}.
  \textbf{Best} metrics are highlighted.}\label{tab:ablation}
\end{table}

To investigate the contribution of each design choice in GraspMF, we perform ablations on the following four components: (i) the auxiliary signed-distance regression objective, (ii) the annealing schedule of the semigroup weight, (iii) the SVD projection for rotation matrices, and (iv) the semigroup loss~\eqref{eq:semi_loss} itself.
Table~\ref{tab:ablation} presents the results for $T = 5$.
We see that removing the auxiliary signed-distance objective (GraspMF-NoSDF, Sec.~\ref{sec:implementation}) causes a drop of $3.8\%$ in ID SR and $5.1\%$ in OOD SR, while increasing EMD in both domains, indicating that grounding the shared feature trunk in object geometry improves performance beyond what the endpoint anchor alone provides.
Holding the semigroup weight constant throughout training (GraspMF-NoSched) costs $2.4\%$ in ID SR and $5.8\%$ in OOD SR, consistent with the annealing rationale of Sec.~\ref{sec:implementation}: the data anchor must dominate before data-free self-consistency is enforced.
Replacing the SVD projection at the neural network's output with the symmetric Gram--Schmidt orthonormalization (GraspMF-GS) leaves ID SR unchanged ($87.51\%$ vs.\ $87.40\%$) and reduces latency ($13.5$ vs.\ $15.5$\,ms), but forfeits $4.7\%$ in OOD SR and increases EMD in both domains.
SVD projection is the optimal solution to the orthogonal Procrustes problem, whereas Gram--Schmidt is a greedy procedure that in general does not minimize the Frobenius distance to the target rotation.
Furthermore, pairing the $9$-D output space with SVD yields smoother gradients, and its redundancy regularizes training, in contrast to pairing a $6$-D output space with Gram--Schmidt~\citep{Zhou2019Continuity}.
Replacing the semigroup loss~\eqref{eq:semi_loss} with the decomposed MeanFlow objective of~\citet{Zhong2026RMFManifolds} (GraspMF-Decomp) lowers ID SR by $25.15$ and OOD SR by $19.44$ percentage points, the largest degradation among the ablated components, which identifies the semigroup consistency formulation as the principal contributor to preserving the implicit contact constraints under a few-step budget.
These results justify the design choices of GraspMF for few-step grasp generation.

\subsection{Robustness to Partial Observations}\label{sec:exp_partial}

We evaluate the robustness of GraspMF and the baseline methods under incomplete object geometry, a common scenario in practical robotic grasping.
For the objects from the ACRONYM dataset, we simulate single-view partial observations by raycasting from a viewpoint drawn at random from a $1000$-point Fibonacci sphere centered at the object's origin, and the returned surface points are subsampled to $N = 1024$ points.
Furthermore, the reference grasp set for this scenario is correspondingly restricted to grasps whose closing volume contains a subset of the sampled geometry.
This view-dependent data sampling protocol is used for training and evaluation (where we average the results over 3 random views per object), for which the results are reported in Table~\ref{tab:partial}.
GraspMF attains the highest OOD SR at $T = 5$, with $68.46\%$, and BRIDG{\scriptsize{E}}R follows at $65.38\%$.
On ID, BRIDG{\scriptsize{E}}R leads with $88.84\%$ SR, while GraspMF at $T = 5$ trails by $2.76\%$ ($86.08\%$ vs.\ $88.84\%$).
GraspMF at $T = 5$ is nonetheless $7.8\times$ faster ($15.5$\,ms vs.\ $120.6$\,ms, Table~\ref{tab:latency_nfe}), and the remaining performance gap can be reduced by drawing larger candidate populations with rejection sampling and replanning loops, which is advantageous in dynamic scenarios.
GraspMF also records the highest OOD EMD ($0.5923$), consistent with the view-dependent restriction of the reference grasp set discussed below: generated grasps covering modes outside the visible-surface reference set raise the measured EMD without necessarily failing in execution.

Absolute performance is lower than in Table~\ref{tab:main} for most methods, owing to three factors intrinsic to the protocol.
First, a single view resolves only the camera-facing surface, so the encoder must infer the occluded geometry to determine contact feasibility, and distinct geometries consistent with the same view render the conditional grasp distribution ambiguous with respect to the observation.
Second, restricting the reference set to grasps contacting the observed surface is a plausible choice, but it reduces the number of supervised grasps per object and makes that set vary with the viewpoint, raising the variance of the endpoint target across training samples.
Third, the normalization frame is the centroid of the observed cloud (Sec.~\ref{sec:implementation}), which is biased toward the visible surface and shifts with the viewpoint, so the translational component of the prior $\rho_0$ is no longer centered on the object.

\begin{table}[t]
  \centering\vspace{2mm}
  \small
  \setlength{\tabcolsep}{3pt}
  \resizebox{\linewidth}{!}{%
  \begin{tabular}{lcccc}
    \toprule
    \multirow{2}{*}{Method} & \multicolumn{2}{c}{SR $(\%)$ $\uparrow$} & \multicolumn{2}{c}{EMD $\downarrow$} \\
    \cmidrule(lr){2-3} \cmidrule(lr){4-5}
    & ID & OOD & ID & OOD \\
    \midrule
    SE3Dif~\citep{Urain2023SE3DiffusionFields}
    & $65.51\stdv{22.97}$ & $56.19\stdv{21.08}$ & $0.4785\stdv{0.1475}$ & $\underline{0.4926}\stdv{0.1083}$ \\
    BRIDG{\scriptsize{E}}R~\citep{Chen2024BRIDGER}
    & $\mathbf{88.84}\stdv{19.60}$ & $\underline{65.38}\stdv{25.25}$ & $0.4621\stdv{0.1121}$ & $0.5721\stdv{0.1312}$ \\
    EGF~\citep{Lim2024EquiGraspFlow}
    & $76.87\stdv{23.06}$ & $47.11\stdv{20.75}$ & $\mathbf{0.3801}\stdv{0.0958}$ & $\mathbf{0.4637}\stdv{0.1051}$ \\
    VSIGD~\citep{Bukhari2025VariationalShapeInference}
    & $49.54\stdv{18.39}$ & $52.18\stdv{18.68}$ & $0.5577\stdv{0.1702}$ & $0.5074\stdv{0.0965}$ \\
    \midrule
    \textbf{GraspMF} ($T = 5$)
    & $\underline{86.08}\stdv{18.23}$ & $\mathbf{68.46}\stdv{23.64}$ & $\underline{0.4030}\stdv{0.1213}$ & $0.5923\stdv{0.1872}$ \\
    \textbf{GraspMF} ($T = 1$)
    & $79.04\stdv{18.57}$ & $63.13\stdv{22.73}$ & $0.4071\stdv{0.1140}$ & $0.5883\stdv{0.1849}$ \\
    \bottomrule
  \end{tabular}%
  }
  \vspace{3mm}
  \caption{
    Grasping performance under single-view partial point-cloud observation.
    The sampling configuration kept the same as reported in Table~\ref{tab:latency_nfe}.
    Metrics are computed for performance over 3 random views per object, where for each view, 100 grasps are generated as before.
    \textbf{Best} and \underline{second-best} metrics are highlighted.
  }\label{tab:partial}
\end{table}

\begin{figure}[t]
  \centering\vspace{-5mm}
  \includegraphics[width=0.75\linewidth]{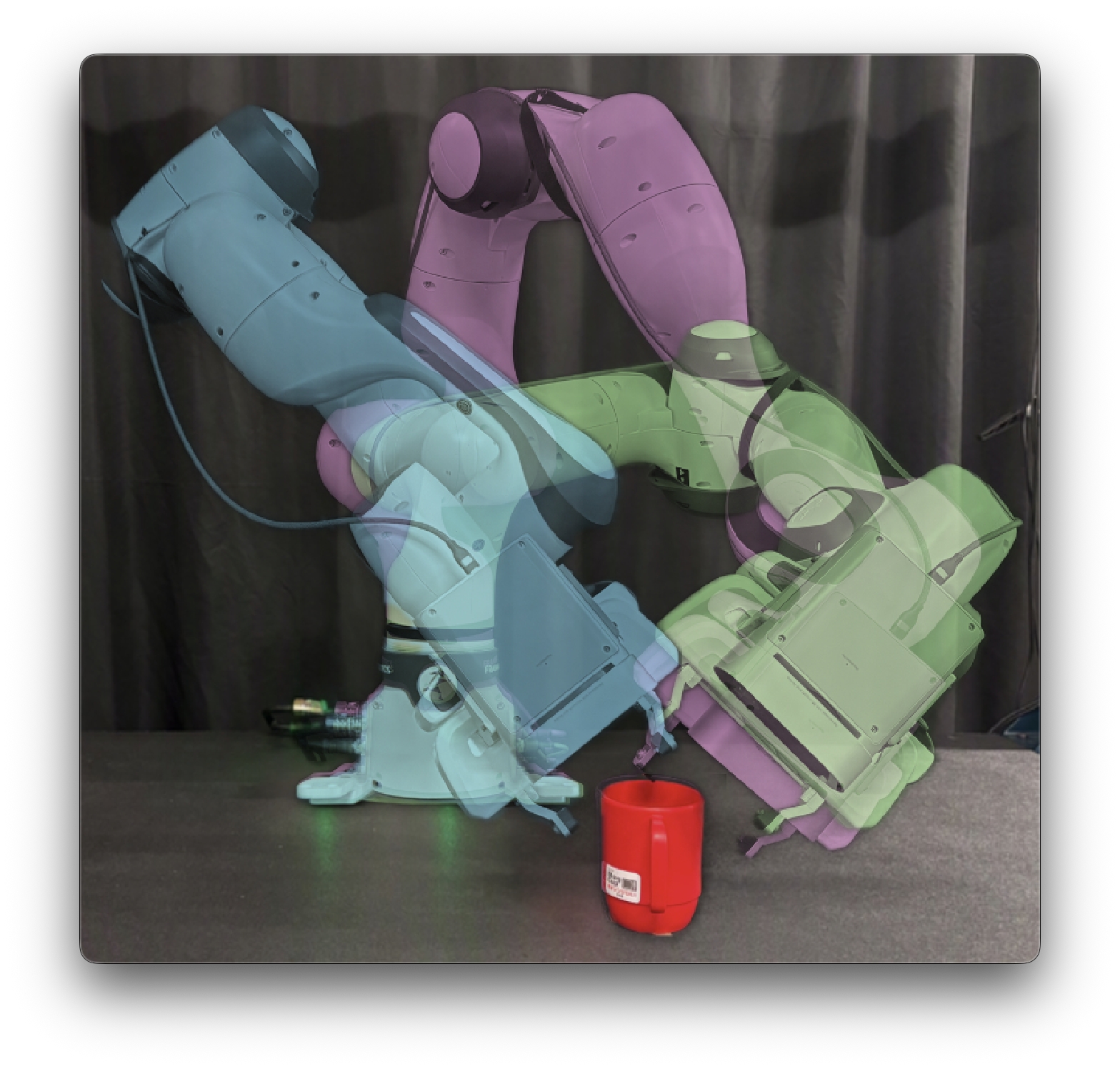}
  \caption{Real-world grasping demonstration on household objects with a wrist-mounted RGB-D camera.
  Three diverse grasps generated by our approach are shown.}\label{fig:three-views}
\end{figure}

\subsection{Real-World Demonstration}\label{sec:exp_real}

To demonstrate the transferability of GraspMF to practical scenarios, we test it on a real-world table-top grasping task.
We use Franka Research 3, a 7-DoF robotic manipulator, equipped with a Franka Hand gripper and a wrist-mounted Orbbec Femto Mega RGB-D camera.
Our evaluation protocol comprises three phases.
First, scene point clouds acquired from the camera are segmented with SAM~\citep{carion2025sam} and fed to GraspMF to generate stable grasp poses.
Then, collision-free trajectories to and from the generated grasp poses are computed using the RRT-Connect implementation from VAMP~\citep{thomason2024vamp}.
Finally, the robot executes the plan to grasp the object, and picks it up vertically to verify grasp stability.
Table~\ref{tab:rw} reports the number of successful grasps out of 10 trials for each object, where GraspMF performs consistently across all objects while operating in the few-step sampling regime ($T \in \{5, 10\}$); Fig.~\ref{fig:three-views} shows representative grasps from these trials.
We complement the results with video demonstrations in the supplementary material.

\begin{table}[t]
  \centering\vspace{2mm}
  \small
  \setlength{\tabcolsep}{4pt}
  \resizebox{0.75\linewidth}{!}{%
  \begin{tabular}{lccc}
    \toprule
    \multirow{2}{*}{Method}
    & \multicolumn{3}{c}{Trials ($\cdot$/10)} \\
    \cmidrule(lr){2-4}
    & Black Mug & Red Mug & Gray Bowl \\
    \midrule
    SE3Dif~\citep{Urain2023SE3DiffusionFields} ($T=70$)
    & $3$ & $6$ & $2$ \\
    BRIDG{\scriptsize{E}}R~\citep{Chen2024BRIDGER} ($T=40$)
    & $\mathbf{10}$ & $8$ & $7$ \\
    EGF~\citep{Lim2024EquiGraspFlow} ($T=20$)
    & $\mathbf{10}$ & $8$ & $\mathbf{10}$ \\
    VSIGD~\citep{Bukhari2025VariationalShapeInference} ($T=70$)
    & $6$ & $9$ & $9$ \\
    \midrule
    \textbf{GraspMF} ($T = 5$)
    & $9$ & $9$ & $\mathbf{10}$ \\
    \textbf{GraspMF} ($T = 10$)
    & $9$ & $\mathbf{10}$ & $\mathbf{10}$ \\
    \bottomrule
  \end{tabular}
  }
  \vspace{1mm}
  \caption{Real-world grasping performance across three household objects.
  Each entry reports the number of successful grasps out of 10 trials.
  \textbf{Best} results for each object are highlighted.}\label{tab:rw}
\end{table}

\section{Conclusion}\label{sec:conclusion}

We presented a Lie-group formulation of MeanFlow for robotic grasp generation in $\SE{3}$.
The construction operates in the left-trivialization of the product Lie group $\cal{G} = \SO{3} \times \R^3$, parameterizes the network by direct prediction of the clean grasp~\eqref{eq:x1_pred}, derives the average velocity~\eqref{eq:trivialized_avg_def} and flow map~\eqref{eq:flow_map} in closed form, and trains through the exact algebraic semigroup identity~\eqref{eq:semigroup_log} composed with a flow-matching endpoint anchor at the time diagonal.
Despite a lightweight neural network architecture, the model supports single- and few-step grasp synthesis at inference, attaining grasp quality comparable to multi-step diffusion and flow baselines on ACRONYM at up to $39\times$ lower latency.
In future work, we plan to explore extensions to general matrix Lie groups relevant in manipulation (e.g., dual-arm $\SE{3} \times \SE{3}$, contact manifolds for non-prehensile interaction) and integration with closed-loop planners that exploit the reduced inference cost.

\bibliographystyle{IEEEtranN}
\bibliography{references} %

\end{document}